\documentclass[lettersize,journal]{IEEEtran}

\usepackage{algpseudocode}
\usepackage{cite}
\usepackage{amsmath,amssymb,amsfonts}
\usepackage{textcomp}
\usepackage{booktabs} 
\usepackage{multirow}
\usepackage{makecell}

\usepackage{subfig}
\usepackage{tabularx}
\usepackage{graphicx}
\usepackage{float}
\usepackage{stfloats}
\usepackage{booktabs}
\usepackage{amsthm}
\usepackage{stfloats}
\usepackage{algorithmicx,algorithm}
\usepackage{graphicx}

\begin{document}

\title{Toward Time-Continuous Data Inference in Sparse Urban CrowdSensing}

\author{Ziyu~Sun, Haoyang~Su, Hanqi~Sun, En~Wang,~\IEEEmembership{Member,~IEEE,} Wenbin~Liu
\thanks{\IEEEcompsocthanksitem Ziyu~Sun, Haoyang~Su, Hanqi~Sun,  En~Wang, Wenbin~Liu are with the College of Computer Science and Technology, Jilin University, Changchun 130012, China, and also with the Key Laboratory of Symbolic Computation and Knowledge Engineering of Ministry of Education, Jilin University, Changchun 130012, China
(e-mail: sunzy2121@mails.jlu.edu.cn;suhy2121@mails.jlu.edu.cn;sunhq5521@mails.\\jlu.edu.cn; wangen@jlu.edu.cn; liuwenbin@jlu.edu.cn).
}
\thanks{(Ziyu Sun and Haoyang Su contributed equally to this work.) (Corresponding author: Wenbin~Liu.)}}



\maketitle

\begin{abstract}
Mobile Crowd Sensing (MCS) is a promising paradigm that leverages mobile users and their smart portable devices to perform various real-world tasks. However, due to budget constraints and the inaccessibility of certain areas, Sparse MCS has emerged as a more practical alternative, collecting data from a limited number of target subareas and utilizing inference algorithms to complete the full sensing map. While existing approaches typically assume a time-discrete setting with data remaining constant within each sensing cycle, this simplification can introduce significant errors, especially when dealing with long cycles, as real-world sensing data often changes continuously. In this paper, we go from fine-grained completion, i.e., the subdivision of sensing cycles into minimal time units, towards a more accurate, time-continuous completion. We first introduce  Deep Matrix Factorization (DMF) as a neural network-enabled framework and enhance it with a Recurrent Neural Network (RNN-DMF) to capture temporal correlations in these finer time slices. To further deal with the  continuous data, we propose TIME-DMF, which captures temporal information across unequal intervals, enabling time-continuous completion. Additionally, we present the Query-Generate (Q-G) strategy within TIME-DMF to model the infinite states of continuous data. Extensive experiments across five types of sensing tasks demonstrate the effectiveness of our models and the advantages of time-continuous completion.
 
\end{abstract}


\section{Introduction}
\IEEEPARstart{W}ith the evolution of information society and the increasing portability of wireless devices, Mobile CrowdSensing (MCS) \cite{2011_mcs_overview, 2023MCS} has recently emerged as a promising paradigm of data collection. Typically, it recruits a large number of users equipped with mobile devices to collect data from specific sensing areas at particular time. Due to budget constraints and presence of unreachable sensing data, traditional MCS can only collect incomplete or even sparse data in most cases. To this end, a modified paradigm called Sparse MCS \cite{2016_sparse_mcs} is proposed, which introduces inference strategies to complete the full sensing data from the partial observations. Sparse MCS has already shown great advantages in some practical applications, such as the air quality monitoring \cite{2020_sparse_air,feng2018mcs}, traffic control \cite{2020_sparse_traffic,ali2021mobile} and urban sensing \cite{calabrese2014urban,liu2019multi}.

\begin{figure}[t]
  \centering
  \includegraphics[width=\linewidth]{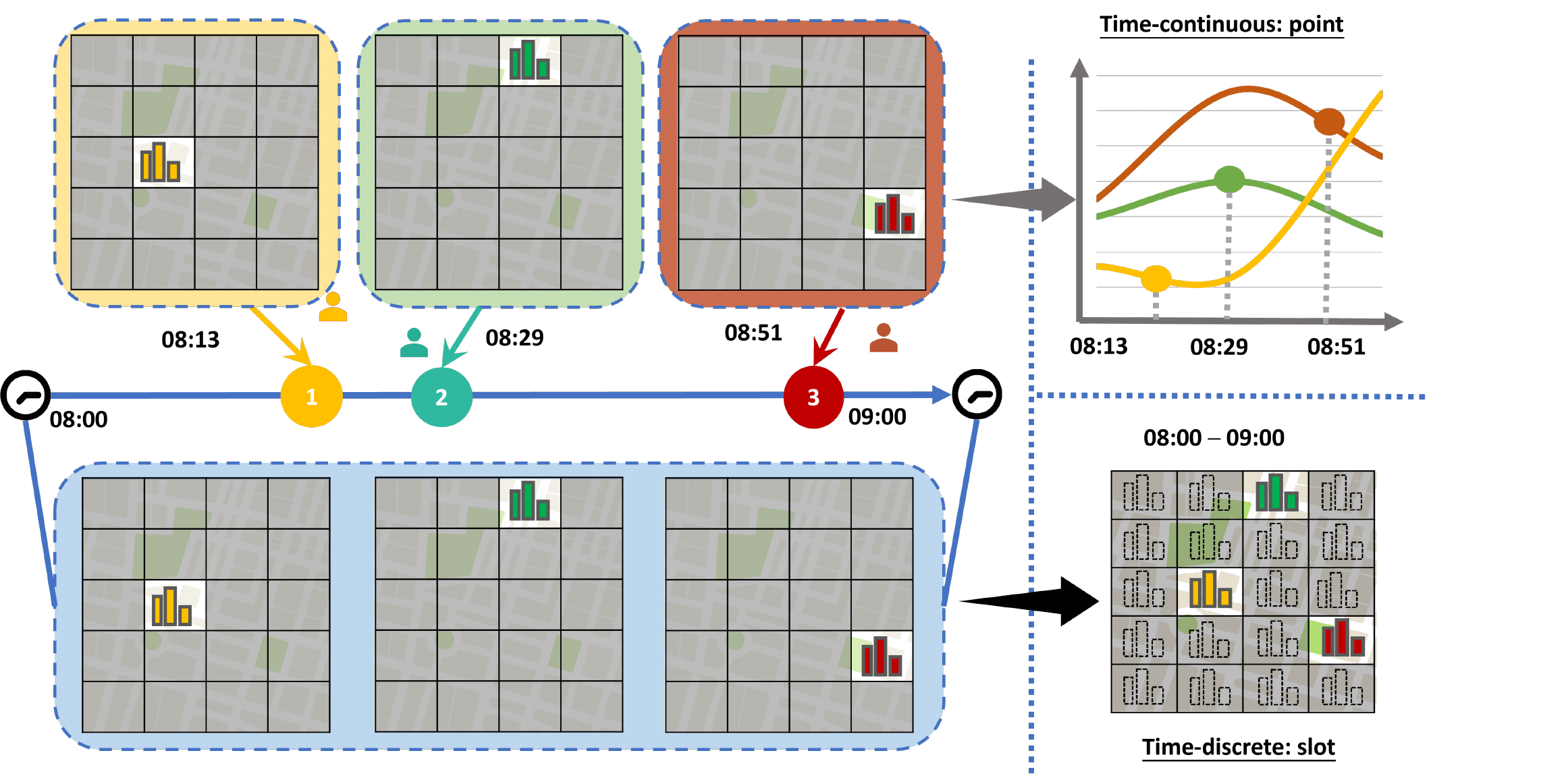}
  \caption{Time-continuous and time-discrete sensing data in Sparse MCS.}
  \label{fig1}
\end{figure}
In Sparse MCS, data inference is the most essential part and has therefore received considerable attention. To reduce cost and simplify inference process, most of existing works study the data inference problem from a time-discrete perspective
\cite{2015_cs_sparsity,2017_framework,2018_cs_variant,2019_framework,liu2022worker,wang2022outlier}. For example, in Fig. \ref{fig1}, a requester would like to analyze urban traffic for a period. Upon receiving the task, existing works typically discretize the time period into units and aggregate the sensed data within each unit. Then, by assuming that the sensed data remains constant within each time unit, they use the data inference methods, such as compressive sensing \cite{2013_cs_traffic,2018_cs_variant} or matrix completion \cite{2018_dmf,2019_gnn} to infer the missing data. However, in practical scenario, the sensing data changes continuously over time. Previous time-discrete approaches may cause significant errors on practical applications that are sensitive to change. For example, temperature or wind speed may fluctuates greatly within a short time due to severe weathers and the rough time-discrete method may fail to capture this dramatic local changes. Therefore, time-continuous data inference has become a crucial issue that urgently needs addressing for Sparse MCS.

In this paper, we adopt a time-continuous perspective, moving away from the traditional method of discretizing time into fixed units. This shift eliminates the assumption that data remains constant within a specific period, making it impossible to aggregate observed data within each time unit to reduce matrix sparsity. Consequently, our first challenge is to handle the \textit{extremely sparse data matrix}. Additionally, in time-continuous scenarios, data is collected in real-time, leading to unequal lengths between sensed data intervals, which affects the relationships between consecutive data points. Thus, our second challenge is to model and utilize these \textit{unequal intervals} effectively, maximizing the temporal information for accurate data inference. Finally, while time-discrete methods could represent the problem with a fixed-size matrix, they cannot infer the continuously changing data at every moment. This leads to our third challenge: how to \textit{complete the data from a continuous perspective}, ensuring comprehensive data inference across all moments.
\begin{figure}[t]
    \centering
    \includegraphics[width=1.0\linewidth]{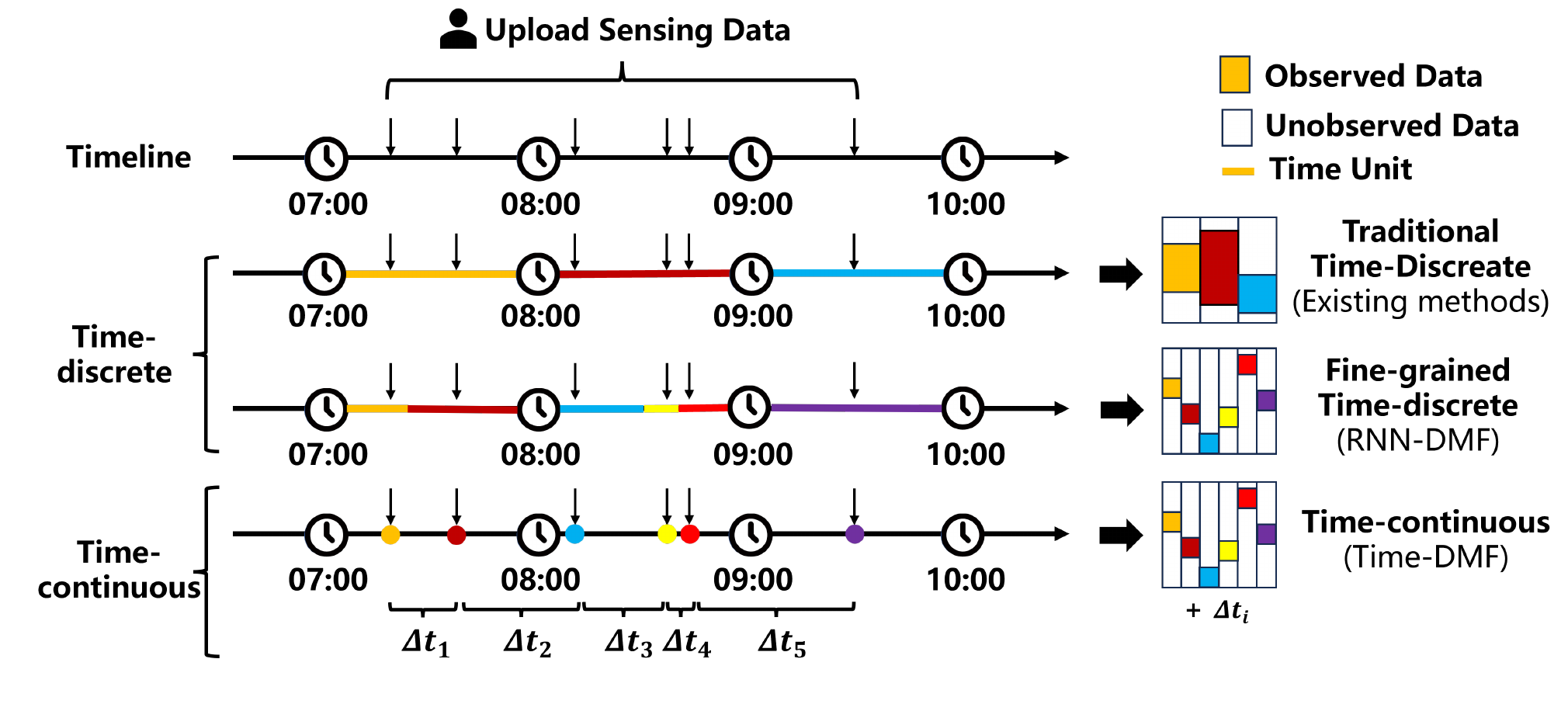}
    \caption{Time-discrete and time-continuous formulation.}
    \label{fig_task_defination}
\end{figure}
To tackle the challenges inherent in time-continuous data completion, we introduce a comprehensive approach that begins by reformulating the problem and progresses to designing a method that addresses these challenges effectively. We start with fine-grained completion, which serves as an intermediary between traditional time-discrete methods and our final goal of continuous completion. Unlike traditional time-discrete completion, which uses predefined unit lengths to discretize the timeline, fine-grained inference divides the timeline based on the actual distribution of submissions, ensuring that each time unit contains only one submission. This adjustment eliminates the need to assume data remains constant within each time unit, thereby offering a more accurate model. However, this approach significantly reduces the data volume in each unit, leading to an extremely sparse spatiotemporal matrix (i.e., our first challenge). To address this, we introduce Recurrent Neural Network-enabled Deep Matrix Factorization (RNN-DMF), a neural network-based framework for data completion that also incorporates a temporal encoder to fully leverage previously hidden information, addressing the challenge of sparsity.

Moving beyond fine-grained completion, we recognize that it still does not fully capture the continuous nature of time. In time-continuous scenarios, we avoid discretizing the timeline entirely and handle each submission directly, preserving the precise arrival times and intervals between submissions—our second challenge. To leverage this additional temporal information, we introduce Time Gates-enhanced Deep Matrix Factorization (TIME-DMF), which further captures the temporal dynamics within intervals through the use of time gates and a more sophisticated propagation pattern. Finally, addressing the challenge of representing the infinite number of moments within a period, we propose the Query-Generate (Q-G) strategy, which works in conjunction with TIME-DMF to model any moment on the timeline, thus providing a comprehensive solution to time-continuous data completion.

Our work has the following contributions:

\begin{itemize}
  \item We reformulate the problem of data inference from a time-continuous perspective. It pays attention to the continuity of data changes and is a much closer approximation of practical MCS problems.
  \item We introduce DMF which is neural network- enabled framework for data completion and extend it to RNN-DMF with a temporal encoder to handle the extreme matrix sparsity in fine-grained completion.
  \item We propose TIME-DMF based on RNN-DMF with time gates to capture temporal information within intervals and its accompanying Q-G strategy which allows users to make queries and dynamically generate responses to achieve time-continuous completion.
  \item Extensive experiments of five types are conducted step by step to validate the effectiveness of our methods.
\end{itemize}

The reminder of this paper is organized as follows. Section II reviews related works. Section III presents the system model and the problem formulation. In Section IV, we introduce the fine-grained completion, which comprises the DMF framework and RNN-DMF model. In Section V, TIME-DMF and its accompanying Q-G strategy for time-continuous completion are discussed. We evaluate the performance of our approaches through extensive experiments in Section VI, followed by the conclusion in Section VII.
\section{Related Work}

\subsection{Sparse MCS}
Mobile CrowdSensing \cite{2011_mcs_overview, 2014_4w1h} is an emerging paradigm that leverages mobile device users to collect data, enabling a wide range of services within the Internet of Things ecosystem \cite{truong2019trust,luo2019improving,sisi2024blockchain}. MCS has been widely used in domains such as traffic supervision\cite{2013_cs_traffic,hu2019uavs}, pollution control \cite{2015_pollution,liu2018third}, and facility management\cite{2016_richmap,suhag2023comprehensive}. Initially, the mainstream algorithms for MCS were based on compressed sensing\cite{2013_cs_traffic} and its various adaptations\cite{2018_cs_variant}. However, as these algorithms were implemented, it became apparent that data collection often exhibited sparsity\cite{2015_cs_sparsity} due to cost constraints and limitations of sensing devices. Consequently, algorithms used for inferring missing data\cite{2012_cs_completion} gained popularity. In 2016, Wang $et\ al.$\cite{2016_sparse_mcs} provided a comprehensive review of MCS methods based on sparse sensed data and systematically introduced frameworks \cite{2017_framework,2019_framework} for data collection and completion. Since then, Sparse MCS has emerged as an evolving paradigm, with many innovative algorithms being developed. Primary areas of work in this field include cell selection\cite{2019rl}, data inference\cite{2018_dmf,2019_gnn} and user privacy protection\cite{2020_privacy}.

In data inference, methods are generally categorized into two classes: dense-supervised \cite{2023itransformer, 2021autoformer} and sparse-supervised \cite{2019_gnn, 2018_dmf, wang2022outlier}. Dense-supervised methods rely on large amounts of complete spatiotemporal data for training. Most Transformer models and their variants, which are powerful in handling time series, fall under this category. In contrast, sparse-supervised methods do not require complete spatiotemporal data for training and instead rely on capturing correlations within sparsely observed data. Despite their differences, neither approach considers the continuous nature of time.

\subsection{Spatiotemporal Granularity}

The goal of sensing technology is to capture more fine-grained spatiotemporal information. Initially, this was achieved by deploying additional sensors\cite{2010_finegrain_sensors_more}  to decrease granularity and improve accuracy at an expensive price. Subsequently, data inference algorithms\cite{2016_finegrain_completion} emerged as a more cost-effective alternative, finding widespread application in various domains such as air monitoring\cite{2017_finegrain_air1,2018_fine_grain_air2}. 

In terms of time granularity, there is comparatively less research on fine-grained or even continuous timeline. Doya $et\ al.$\cite{2000_continuous_rf} investigated reinforcement learning in continuous spatiotemporal domains, while Brockwell $et\ al.$\cite{2001_continuous_theory} explored improvements to the ARMA algorithm on a continuous timeline. Higuchi $et\ al.$\cite{1988_uneven1} and Kidger $et\ al.$\cite{2020_uneven2} studied cases where events are unevenly distributed over time. In practical applications, Zhu $et\ al.$\cite{2017_time_lstm} examined the impact of non-uniform distribution of customer consumption from a time-continuous perspective in recommendation systems, introducing time gates into deep learning models for handling unequal time intervals between events. Drawing from these insights, in this paper, we conduct a comprehensive analysis on reducing the temporal granularity of sensing tasks and aim to achieve time-continuous completion in Sparse MCS. 

\section{System Model and Problem Formulation}

\subsection{System Model}
Fig. \ref{fig1} and Fig. \ref{fig_task_defination} showcase different methods for formulating real-world spatiotemporal data. The majority of existing research has relied on the time-discrete approach depicted in the lower half of Fig. \ref{fig1}, which serves as a coarse-grained approximation of the real world. In this section, we will begin with the time-discrete formulation and subsequently introduce our novel time-continuous approach to provide a deeper understanding of our concept.

In Sparse MCS tasks, our sensing map is divided into N sub-regions. At moment $t$, a worker submits data of the $n-th$ sub-region. We use a position vector $c_{(t)} \in R^N$ and a data vector $y_{(t)}^\prime \in R^N$ to represent the submission. The position vector $c_{(t)} = [0, 0, \dots, 1,0\dots 0]$ indicates the index of the subarea submitted by the worker. If the submitted data is from the $i-$th sub-region, the $i-$th element of $c_{(t)}$ is set to $1$, and all other elements are set to $0$. The data vector $y_{(t)}^\prime$ indicates the value of the submitted subarea. The $i-$th element of $y_{(i)}^{\prime}$ is the submitted value, and other elements of $y_{(i)}^{\prime}$ are set to meaningless values like 0 or negative numbers. 

Assuming that there are $M$ submissions on the entire timeline, the results of $M$ submissions are organized into position matrix and data matrix, we have $C\in R^{N\times M}$and $Y' \in R^{N\times M}$ by stacking all the submissions together:
{\setlength\abovedisplayskip{5pt}
\setlength\belowdisplayskip{5pt}
\begin{align}
C &= [c_{1}^{ \mathrm{T}}, c_{2}^{ \mathrm{T}},\cdots, c_{M}^{ \mathrm{T}}], \\
Y^{\prime} &= [y_{1}^{\prime \mathrm{T}}, y_{2}^{ \prime \mathrm{T}},\cdots, y_{M}^{\prime \mathrm{T}}].
\label{eq_CY'}
\end{align}}%
Similarly, $Y = [y_{1}^{ \mathrm{T}}, y_{2}^{ \mathrm{T}},\cdots, y_{M}^{ \mathrm{T}}]$ represents the real values of each sub-region at each submission time. So we have:
{\setlength\abovedisplayskip{5pt}
\setlength\belowdisplayskip{5pt}
\begin{align}
  Y ^{\prime}= Y \odot C,
\label{eq_1}
\end{align}}%
where $\odot$ represents the Hadamard product.

In traditional Sparse MCS tasks, the timeline is evenly discretized into time units of predefined length, and we no more distinguish the difference of arriving time of submissions within the same time unit. Suppose that we manually discretize the given timeline into $P$ time units, we can have our new matrices $C^{(D)}\in R^{N \times P}$ and $Y^{(D) \prime} \in R^{N \times P}$ by putting each submission into the time unit they locate in and merge all the submissions within the same unit.
\begin{figure}[t]
\centering
\includegraphics[width=1.1\linewidth]{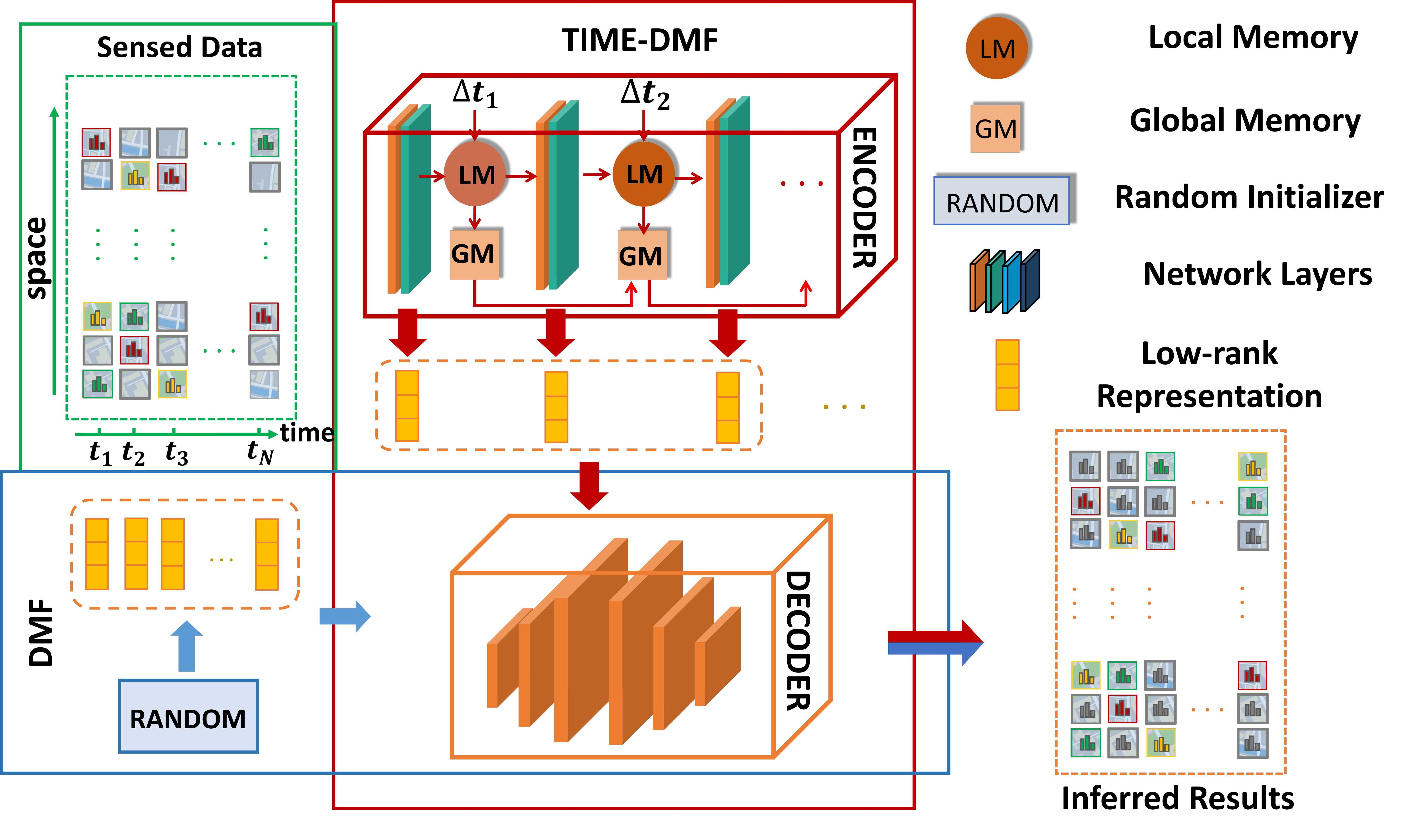}
\caption{Relationship between DMF and TIME-DMF.}
		\label{fig2}
\end{figure}
{\setlength\abovedisplayskip{5pt}
\setlength\belowdisplayskip{5pt}
\begin{align}
C^{(D)} &= [c_{1}^{(D) \mathrm{T}}, c_{2}^{(D) \mathrm{T}},\cdots, c_{P}^{(D) \mathrm{T}}], \\ 
Y^{(D)\prime} &= [y_{1}^{(D) \prime \mathrm{T}}, c_{2}^{(D) \prime  \mathrm{T}},\cdots, c_{P}^{(D)\prime  \mathrm{T}}], \\ 
c_{p}^{(D)}, y_{p}^{(D) \prime} &= merge(c_{p_0}, c_{p_1}, \cdots, y_{p_0}^{\prime}, y_{p_1}^{\prime}, \cdots),
\label{eq_CD_YD}
\end{align}}%
where $c_{p_i}$ represents the $i-th$ submission arrived within the $p-th$ time unit and $merge(\cdot)$ represents the merging operation. Technically, for each subarea, if there is a submission within the given $p-th$ time unit, the corresponding position in $c^{(D)}_{p}$ will be set to one and the submitted value will be filled in $y^{(D) \prime}_{p}$ . If there are multiple values observed for one location, the final value would be set to the average. By the merging operation, the sparsity of  $C^{(D)}$ will be much less than that of matrix $C$. After the discretization and merging operation, we no longer keep the specific arrival time of all submissions and their intervals but use $p-index$ to represent their arrival time. Then our task is to infer the missing values in $Y^{(D)\prime}$ to know the value of each location on the map at each time unit.
{\setlength\abovedisplayskip{5pt}
\setlength\belowdisplayskip{5pt}
\begin{align}
\hat Y = f(Y^{(D)\prime}).
\label{eq3}
\end{align}}%
However, it is obvious that some temporal information has been ignored during the discretization and merging process such as the precise arrival time and the intervals. Inspired by this observation, we propose a new system model to reduce information loss during problem modeling phase. For the $M$ submissions, we use an additional vector $T = [t_{(1)}, t_{(2)}, \cdots, t_{(M)}]$ to represent the accurate arrival time of each submission and no longer discretize $C$ and $Y^{\prime}$ to $C^{(D)}$ and $Y^{(D) \prime}$. What's more, since in time-continuous completion each column in the observed matrix represents a specific moment instead of a period, we not only want to complete the matrix $Y^{\prime}$, but also the infinitely many moments that do not exist in $Y^{\prime}$.  In order to achieve this, we divide the challenging time-continuous completion task into to two subtasks.

Our first task is to achieve a mapping $f(\cdot)$ only to complete the matrix $Y^{\prime}$ which fully utilizes the temporal information in $T$. This is similar to the time-discrete scenarios.
{\setlength\abovedisplayskip{5pt}
\setlength\belowdisplayskip{5pt}
\begin{align}
\hat Y = f(Y^{\prime}).
\label{eq3.C}
\end{align}}%
After that, we want to provide accurate inference for any moment on the time line. Obviously there are countless values distributed along the time line, and our solution is to find a model that can provide accurate responses $\hat y \in R^N$ for any given time moment $t$.
{\setlength\abovedisplayskip{5pt}
\setlength\belowdisplayskip{5pt}
\begin{align}
\hat y = g(Y^{\prime}, t), t \in (t_{0}, t_{M}).
\label{eq3}
\end{align}}%

\subsection{Problem Formulation}
\textbf{Problem} [Completion and Generation on Continuous Timeline]: Given sparse sensed data $Y^\prime$ and time vector $T$, we aim to achieve the following two objectives:
\begin{itemize}
\item  Identify a mapping $f(\cdot)$ to complete all the unsensed data in the matrix $Y^\prime$. The mapping $f(\cdot)$ should adequately consider the high sparsity of $Y^\prime$ and the temporal information in $T$.
\item Identify a model $G(\cdot)$ to accomplish the completion at any given time $t^\prime$. $y^{(t^\prime)\prime}$ can be a column in $Y^{\prime}$ or not.
\end{itemize}

In this process, the mean square error is used to measure the quality of the completed and generated data. The following value should be minimized:
{\setlength\abovedisplayskip{5pt}
\setlength\belowdisplayskip{5pt}
\begin{align}
\epsilon (Y, Y^\prime) = \sum^{N}_{i} \sum^M_j \vert Y_{ij} - Y^{\prime}_{ij}\vert.  
\label{eq4}
\end{align}}%

\section{Fine-grained Completion with RNN-DMF}
    
Fine-grained completion is the first step towards time-continuous completion. With the insight that we can construct our spatiotemporal matrix by including each submission in a unique time unit, we could align with the previous time-discrete problem setting but eliminate the assumption that data stays constant within each time unit.  We propose Deep Matrix Factorization (DMF) \cite{2018_dmf} as a foundational framework for the following works. Due to its neural network-enabled structure, it can be updated by adding modules of different functions. In fine-grained completion, there is one submission within each time unit so that we only know the information of one spatial location, leading to the great reduction of spatial information within a unit and the extreme sparsity of observation matrix. In order to solve this problem, we further propose Recurrent Neural Network-enabled Deep Matrix Factorization (RNN-DMF) by introducing a temporal encoder into DMF. The encoder is able to utilize temporal information to compensate for the loss of spatial information.

\subsection{Deep Matrix Factorization (DMF)}
Given a sparse matrix $Y^{\prime}_{m \times n}$, the traditional method is to take full rank decomposition of the estimated matrix $\hat{Y}_{m \times n}$ by using the property that the real matrix $Y _{m \times n}$ has the lower rank and can be decomposed. Assuming that  $rank(Y) = r$, then the full rank decomposition can be expressed as:
{\setlength\abovedisplayskip{5pt}
\setlength\belowdisplayskip{5pt}
\begin{align}
Y _{m \times n} = P _{m \times r} Z _{r \times n},
\label{eq5}
\end{align}}%
where P  is a full rank matrix and Z is a row full rank matrix. Therefore, any column $y_t$ of $Y_{m \times n}$ can be expressed as $y_t = Pz_t$. In this formula, $z_t$ represents the low-rank vector that fully contains the information of the $t-$th column of $Y_{m \times n}$, which is also the $t-$th moment in real scenarios. $P$ denotes the projection from the low-rank vectors to the inferred matrix. Note that Eq. $\eqref{eq5}$ assumes that the spatiotemporal correlations between data are linear. To model the widely existing nonlinear spatiotemporal correlations, DMF was proposed. Similarly, we use the function $f(\cdot)$ to represent the nonlinear correlations between space and time.
{\setlength\abovedisplayskip{5pt}
\setlength\belowdisplayskip{5pt}
\begin{align}
y = f(z), Y = f(Z).
\label{eq6}
\end{align}}%
We aim to obtain suitable $z$ and $f(\cdot)$ by fitting them with a deep neural network (DNN). Specifically, assuming the neural network consists of K hidden layers, and their parameters are
{\setlength\abovedisplayskip{5pt}
\setlength\belowdisplayskip{5pt}
\begin{align}
 W^* &\triangleq \{W^{(1)}, W^{(2)}, \cdots, W^{(K)}, W^{(K+1)}\}, \\
 b^* &\triangleq \{b^{(1)}, b^{(2)}, \cdots,b^{(K)}, b^{(K+1)} \}.
\label{eq7}
\end{align}}%
The corresponding activation function set is
{\setlength\abovedisplayskip{5pt}
\setlength\belowdisplayskip{5pt}
\begin{align}
g^* \triangleq \{ g^{(1)}(\cdot), g^{(2)}(\cdot), \cdots, g^{(K)}(\cdot),g^{(K+1)}(\cdot) \}.
\label{eq8}
\end{align}}%
In the set $W^*$, $b^*$, $g^*$, the $(K+1)th$ term represents the parameter or the activation function from the hidden layer K to its output layer. The nonlinear function is expressed as:
\begin{equation}
\begin{aligned}
f(z) = &g^{(K+1)}(W^{(K+1)}g^{(K)}(W^{(K)},\dots, \\
g^{(1)}(&W^{(1)}z + b^{(1)}) \dots + b^{(K)})+b^{(K+1)}).
\end{aligned}
\label{eq9}
\end{equation}
Different from traditional neural networks, the input $z$ and the neural network parameters are both optimizable parameters in this context. The optimization object is as follows:
{\setlength\abovedisplayskip{5pt}
\setlength\belowdisplayskip{5pt}
\begin{align}
min\ \frac{1}{2n} \Vert (Y^{\prime}-f(Z)) * C \Vert.
\label{eq10}
\end{align}}%
In this way, the inferred results can be obtained:
{\setlength\abovedisplayskip{5pt}
\setlength\belowdisplayskip{5pt}
\begin{align}
\hat{Y} = f(Z).
\label{eq11}
\end{align}}%

\begin{figure}[t]
\centering
\includegraphics[width=1\linewidth]{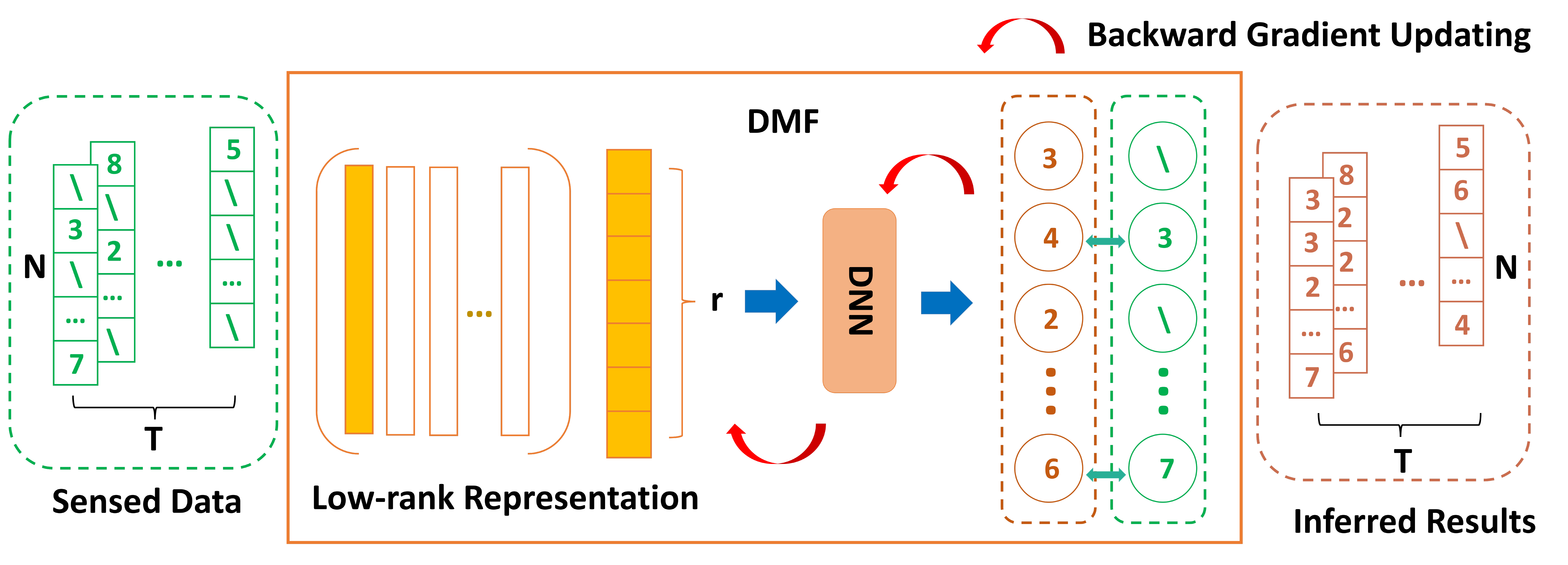}
\caption{The structure of DMF.}
\label{fig3}
\end{figure}

The detailed structure of DMF is shown in Fig. \ref{fig3}. DMF is a neural network framework that is already  able to handle basic data completion tasks. However, it is not tailored for any specific scenarios. By inserting DNN modules of different functions into DMF, we can further enable DMF to deal with data of various specific properties.

\subsection{Recurrent Neural Network-enabled Deep Matrix Factorization (RNN-DMF)}   
 In each time step, the proposed DMF framework has low-rank vectors that can be concatenated to form the complete low-rank representation of spatiotemporal data. While in the training process, each low-rank vector is learned separately and is not visible to each other. This limits the ability of DMF to fully capture temporal information. However, capturing and utilizing temporal information is an urgent in fine-grained completion due to the extreme sparsity of matrices. For this reason, we further propose Recurrent Neural Network-enabled Deep Matrix Factorization
(RNN-DMF).

\begin{figure}[t]
\centering
\includegraphics[width=1.0\linewidth]{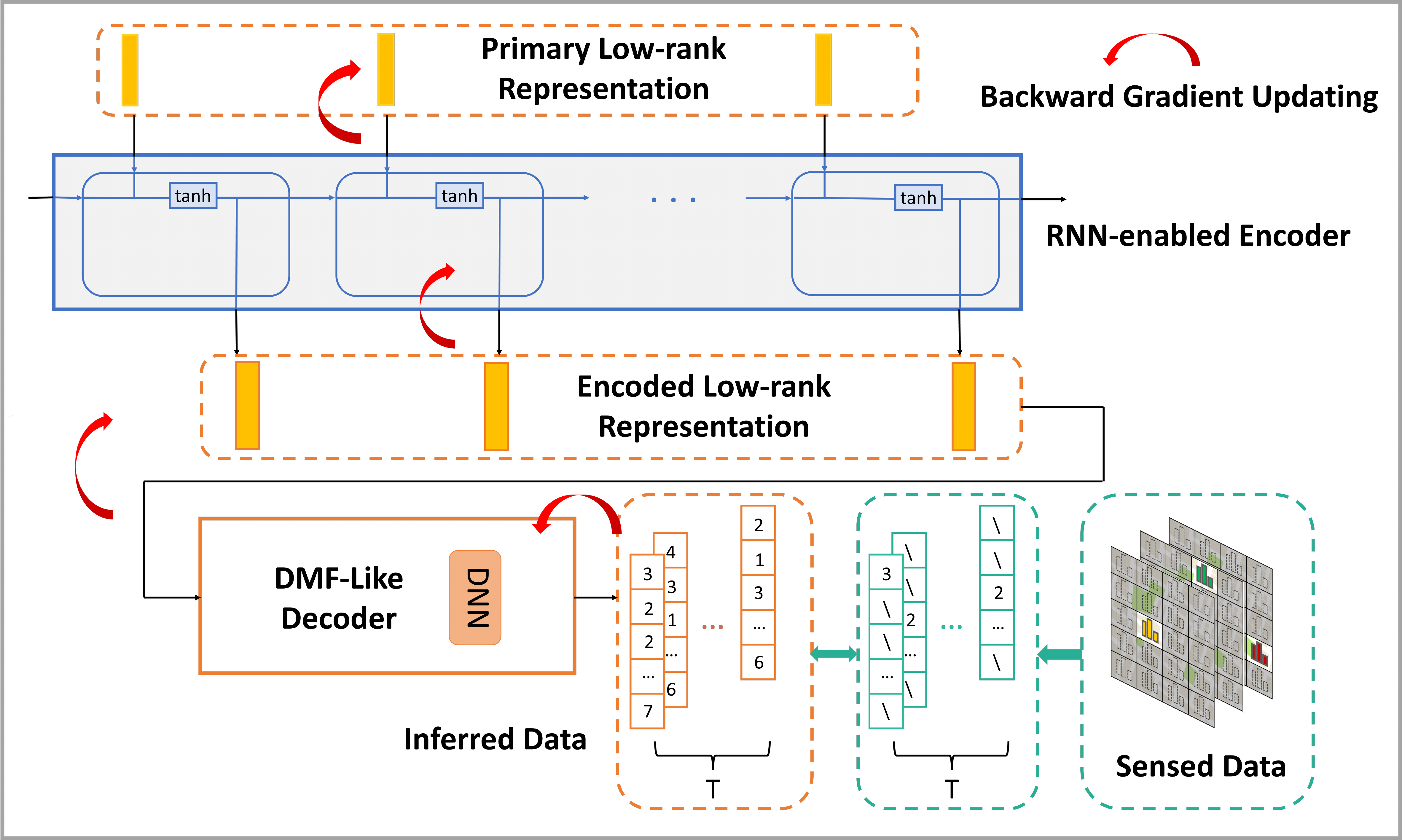}
\caption{The inner structure of RNN-DMF.}
\label{fig4}
\end{figure}

\begin{figure}[tbp]
\centering
\includegraphics[width=1.0\linewidth]{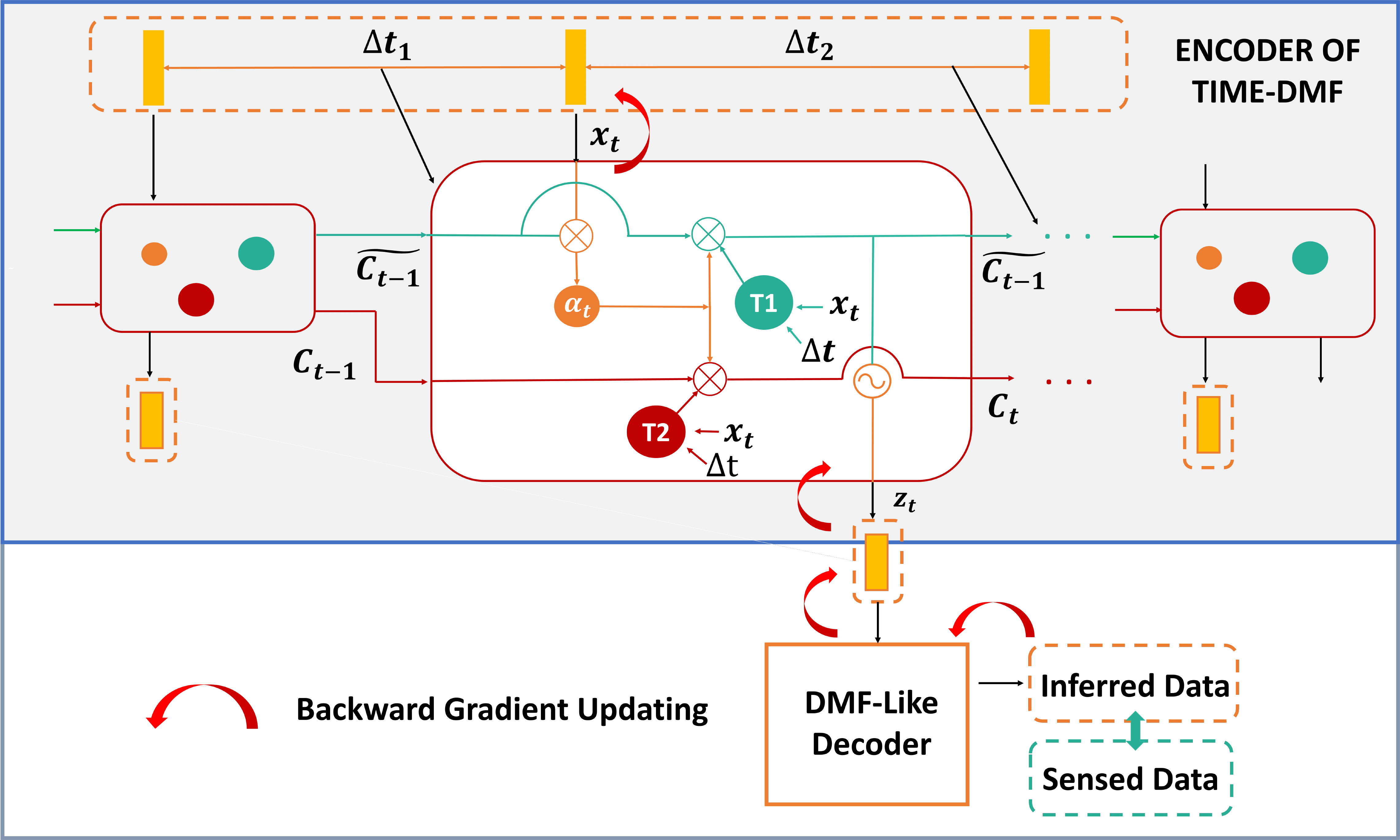}
\caption{The inner structure of TIME-DMF.}
\label{fig5}
\end{figure}

Fig. \ref{fig5} shows that RNN-DMF is composed of two key modules: an upstream RNN-enabled encoder and a downstream DMF-like decoder. Unlike DMF, which initializes low-rank representations randomly, RNN-DMF accounts for relationships between low-rank vectors. The RNN structure enables the encoder to generate low-rank representations by incorporating temporal correlations, sharing parameters \(U\), \(W\), and \(V\) across time steps. The hidden state \(S_t\) is generated based on the previous state \(S_{t-1}\) and the primary low-rank vector \(X_t\) at each timestamp.
{\setlength\abovedisplayskip{5pt}
\setlength\belowdisplayskip{5pt}
\begin{align}
S_t = f(U \cdot X_t + W \cdot S_{t-1}).
\label{eq12}
\end{align}}%
As DMF, $S_0$ and the primary low-rank vector of each step is randomly initialized and serves as optimizable parameters during training. The encoded low-rank representation is then generated by a final projection. This process is similar to what traditional RNN does.
{\setlength\abovedisplayskip{5pt}
\setlength\belowdisplayskip{5pt}
\begin{align}
z_t= g(V \cdot S_t).
\label{eq13}
\end{align}}%
At this step, we finally have our encoded low-rank vectors with temporal information integrated. They will then be concatenated and decoded, serving as the final completion results. This is done by the downstream DMF-like decoder:
{\setlength\abovedisplayskip{5pt}
\setlength\belowdisplayskip{5pt}
\begin{align}
Z &= [z_{1}^{\mathrm{T}}, z_{2}^{\mathrm{T}}, \cdots, z_{M}^{\mathrm{T}}], \\
\hat{Y} &= f(Z).
\label{eq14}
\end{align}}%
It is obvious that RNN-DMF performs strictly superior to DMF theoretically.  This is because DMF generates its low-rank vectors randomly and independently without considering temporal correlations. RNN-DMF considers the possible temporal correlations between low-rank vectors during the generating process, which is at least better than complete random. When sensed data is extremely sparse, there is very limited information for use within each time step, making it urgently necessary to share information between time steps. This explains why RNN-DMF performs significantly better than DMF, especially on extremely sparse matrices.

\section{Time-continuous Completion with TIME-DMF}

In the previous fine-grained completion, we address the challenges of the reduction of spatial information and the extreme sparsity of observation matrices. Based on that, we can finally introduce the time-continuous completion. In traditional works and fine-grained completion, the intervals between submissions are ignored during the discretization stage. However, in real-life scenarios, users submit data at real time and the interval between submissions are of great importance for inference. This oversimplification is also the reason why traditional time-discrete completion methods cannot fully capture temporal information.

Intuitively, it is not the absolute arrival time of submissions but their intervals that matter. The sensing data with a long period of time interval will vary significantly but the sensing data at adjacent time may be quite similar. This is the intuition of Time Gates-enhanced Deep Matrix Factorization (TIME-DMF) for our introducing time gates and a more complex propagation pattern to model the influence of intervals on correlations between time steps. Furthermore, in fine-grained scenarios, period is sliced into a finite number of units and the observation matrix is of fixed size. But from a time-continuous perspective, data changes continuously and there are infinitely many states of sensed data in a given period. In order to characterize data of all states at an acceptable price, we propose Q-G strategy. It allows users to query data of any time and leverages the generative ability of TIME-DMF to dynamically respond. By combining TIME-DMF and the Q-G strategy, we achieve the ultimate time-continuous completion. In this section, we will introduce the details of TIME-DMF and the Q-G strategy. Finally, we will conclude the complete flow of TIME-DMF algorithm for time-continuous data completion tasks. 
\subsection{Time Gates in TIME-DMF}
The inner structure of TIME-DMF is shown in Fig. \ref{fig5}. Inspired by \cite{2017_time_lstm}, we design two types time gates to manage a more complex pattern of information propagation. The first time gate controls global memory. All data within the observed time period should follow a certain underlying distribution, reflecting the overall characteristics of the dataset, such as the mean or periodicity in real-life scenarios. The second time gate controls local memory. The distribution of data at adjacent time shows the local increase/decrease or other short-term trends. We combine these two memories to ensure the completion result is rational both globally and locally. 

The two kinds of memories are controlled by different parameters and are updated independently. Intuitively, intervals determine the correlations between time steps. The shorter the interval between two time steps, the more similar they are expected to be. So we utilize intervals to control the update of memories. If the memories are slightly updated between the two adjacent time steps, their completion results will be very similar. Conversely, if the memories have a substantial update, the following results will greatly defer from the previous ones. We use the interval as a parameter within each time gates.
{\setlength\abovedisplayskip{5pt}
\setlength\belowdisplayskip{5pt}
\begin{align}
T1_m &= \sigma(x_m W_{x1} + \sigma_{\Delta t}(\Delta t_m W_{t1}) + b_1), \\
T2_m &= \sigma(x_m W_{x2} + \sigma_{\Delta t}(\Delta t_mW_{t2}) + b_2).
\label{eq15}
\end{align}}%
\begin{algorithm}[t]
\caption{Deep Matrix Factorization with Time Gates (TIME-DMF)}\label{alg}
\begin{algorithmic}[1]
\Require $C_{N \times M} $, $Y^{\prime}_{N \times M}$, $t$, $T = [t_{1}, t_{2},\cdots, t_{M}]$,\ $x = [x_{1}, x_{2}, \cdots, x_{M}]$
\Ensure $\hat{Y}$
\State Random Init $x$;
\If{$t \notin T$ }
    \State $T \gets [t_{1}, t_{2},\cdots, t_{k}, t, t_{k+1},\cdots, t_{M}]$;
    \State $x \gets [x_{1}, x_{2}, \cdots,x_{k}, x, x_{k+1}, \cdots, x_{M}]$;
    \State $C_{N \times M} \gets C_{N\times (M+1)}$, $Y^\prime_{N \times M} \gets Y^\prime_{N\times (M+1)}$;
\EndIf
\While{not convergent}
	\For{$x^{(1)}$ to $x^{(M)}$}
            \State $z_t,  \tilde{C}_t, C_t \gets encoder(x^{(t)}, \tilde{C}_{t-1}, C_{t-1})$;
        \EndFor
\State $\hat{Y} \gets decoder([z_1^{\mathrm{T}}, z_2^{\mathrm{T}}, \cdots, z_M^{\mathrm{T}}])$;
\State calculate and reduce $ \Vert (Y^\prime-\hat{Y}) * C \Vert$;
\EndWhile 
\end{algorithmic}
\end{algorithm}
The function of two memories is similar to that of the hidden state in RNN, which is shown in Eq. $\eqref{eq12}$.

In TIME-DMF, the global memories and local memories operate in different manners. The local memory has a more direct impact on the completion of each step:
{\setlength\abovedisplayskip{5pt}
\setlength\belowdisplayskip{5pt}
\begin{align}
    a_t = f(U \cdot X_t + W \cdot \tilde{C}_{t-1}).
\label{eq17}
\end{align}}%
The global memory flows more consistently and has a broad effect on the update of memories. We use the two time gates calculated formerly to control the update of the two memories. In this way we make use of the information within intervals.
{\setlength\abovedisplayskip{5pt}
\setlength\belowdisplayskip{5pt}
\begin{align}
\tilde{C}_t &= f[C_t \odot T1_m + a_t \odot (1-T1_m)], \\
{C_t} &= f[C_t \odot T2_m + a_t \odot (1-T2_m)].
\label{eq18}
\end{align}}%
Lastly, as the local memory can better assess the data status at the current moment, it is converted into an low-rank vector representing time $t$ through the output gate.
{\setlength\abovedisplayskip{5pt}
\setlength\belowdisplayskip{5pt}
\begin{align}
z_t = f(V \cdot \tilde{C_t}).
\label{eq18}
\end{align}}%
At this point, the upstream encoder has been obtained. The encoder is capable of doing time-continuous completion as it can deal with submissions at any given moments. We further use the DMF-like decoder to map the low-rank representation to yield the final completion results. Similar to RNN-DMF, TIME-DMF adopts a joint training method for the upstream and downstream networks.
\subsection{Q-G Strategy}

Given a sparse observation matrix, RNN-DMF or TIME-DMF can infer the missing data. However, unlike previous methods relying on discrete slicing, in time-continuous scenarios, there are infinitely many possible moments on the timeline. Thus, it is impossible for the prior Submit-Complete strategy to infer data at all moments. To overcome this challenge, we propose the Query-Generate strategy which allows users to query data for any moment within the given period and dynamically generate the results. To use TIME-DMF for data generation, we leverage its property of completing data by first generating low-rank representation. We can insert a randomly initialized low-rank vector into the previous low-rank representation and then proceed with the usual completion process. The only difference is that the inserted vector does not affect the backward gradient propagation process.

\subsection{The Complete Flow of TIME-DMF}
 For a time query provided by the user, we first check if there is an existing column in the given sparse matrix. If not, we insert an empty column into matrix  $Y^\prime$ and update the intervals of its both sides. Concurrently, we insert a column with all zeros into matrix C to ensure the query won't affect existing completion process. We then deploy our encoder and decoder sequentially to complete the data matrix $Y^\prime$ and minimize the loss between the completed data and known data at observation positions. In this process, the encoder, decoder and the completed results are continuously updated by calculating and reducing the optimization objective.  The complete algorithm flow is shown in Alg. \ref{alg}.

\section{Experimental Validation}

In this section, we first introduce the datasets and the comparison methods. Then we present performance evaluation of our proposed results. In particular, our experiments can be divided into the answers to the following research questions:

\begin{itemize}
    \item \textbf{RQ1}: Does our RNN structure offer better completion performance for extremely sparse matrices?
    \item \textbf{RQ2}: Do our time gates truly leverage the time interval?
    \item \textbf{RQ3}: In time-continuous, the model needs to have additional generative capabilities. How effective is the generative capability of TIME-DMF?
    \item \textbf{RQ4}: Even our method achieves time-continuous completion, is time-continuous completion truly more effective than time-discrete completion?
    \item \textbf{RQ5}: In the domain of spatiotemporal data, transformers seem to have become the mainstream approach. Why do we choose not to use transformer architectures?
\end{itemize}

\begin{table*}[]
\centering
\caption{Statistics of four evaluation datasets}
\begin{tabular}{@{}ccccc@{}}
\toprule
 & U-Air & Sensor-Scope & TaxiSpeed & Highway England \\ \midrule
\multicolumn{1}{c}{City/Country} & \multicolumn{1}{c}{Beijing (China)} & \multicolumn{1}{c}{Lausanne (Switzerland)} & \multicolumn{1}{c}{Beijing (China)} & \multicolumn{1}{c}{England} \\ \midrule
\multicolumn{1}{c}{Data} & \multicolumn{1}{c}{PM2.5} & \multicolumn{1}{c}{Humidity} & \multicolumn{1}{c}{Traffic speed} & \multicolumn{1}{c}{People counting} \\ \midrule
\multicolumn{1}{c}{Subarea} & \multicolumn{1}{c}{36 subareas} & \multicolumn{1}{c}{57 subareas} & \multicolumn{1}{c}{100 road segments} & \multicolumn{1}{c}{15$\sim$25 subareas } \\ \midrule
\multicolumn{1}{c}{Cycle \& Duration} & \multicolumn{1}{c}{1h \& 11d} & \multicolumn{1}{c}{0.5h \& 7d} & \multicolumn{1}{c}{1h \& 4d} & \multicolumn{1}{c}{0.25h \& 0.5$\sim$3months} \\ \midrule
Mean ± Std. & 79.11 ± 81.21 & 84.52 ± 6.32 & 13.01 ± 6.97 & 112.42 ± 30.54 \\ \bottomrule
\end{tabular}
\end{table*}

\subsection{Datasets}

\begin{itemize}
  \item  \textbf{U-Air} \cite{2013UAIR}  is utilized to gather significant air quality data, specifically PM2.5 and PM10 levels, via monitoring stations located in Beijing, China.
  \item \textbf{Sensor-Scope} \cite{2010sensorscope} is employed to collect a diverse array of environmental readings through the deployment of numerous static sensors on the EPFL campus. A representative type of sensing, namely humidity, is selected for evaluation purposes.
  \item \textbf{TaxiSpeed} \cite{2014taxispeed} gathers traffic speed data pertaining to road segments in Beijing, China by utilizing GPS devices installed on taxis.
  \item \textbf{Highways England (HE)} \cite{datasetHE} serves as a resource in providing information pertaining to travel times, traffic flow rates, incidents, event data and camera imagery for England's major motorways. Due to its large space and time range, we manually selected spatiotemporal data from multiple adjacent regions and adjacent time periods, thus the data size is larger and more flexible.
\end{itemize}

\subsection{Comparison Methods}

\subsubsection{Comparative models for completion task}

\par Sparse-supervised methods which only rely on sparse observed data for training:

\begin{itemize}
    \item \textbf{MC}, a classic linear matrix completion method, assumes a linear relationship $Y = PZ$.
    \item \textbf{KNN-S}, a variant of the K-Nearest Neighbors algorithm. KNN-S retrieves information from the K closest sub-regions to the region to be imputed and uses their average value as the imputation result.
    \item \textbf{GP} algorithm, a method that assumes the spatial distribution of data in the same cycle obeys the Gaussian distribution. The unknown data are inferred by calculating the expectation and variance of the known data.
    \item \textbf{DMF}, which has been introduced previously and also serves as a basic component of our method.
    \item \textbf{STformer}\cite{2023STformer}, a transformer-based model with multiple designed embedding and attention layers to capture spatiotemporal relationship. STformer is specially designed to be trained with only sparse observed data.
\end{itemize}
    
Dense-supervised methods which rely on complete observed data to train the model:

\begin{itemize}
    \item \textbf{iTransforme}r\cite{2023itransformer}, a variant which applies the attention and feed-forward network on the inverted dimensions. This is fine-tuned for completion tasks.
    \item \textbf{AutoFormer}\cite{2021autoformer}, another variant of transformer which entangles different blocks in the same layers during supernet training.It is also fine-tuned for completion tasks.
\end{itemize}

\subsubsection{Comparative Predictive models for generative task}

\begin{itemize}
  \item \textbf{LINEAR}, which applies the linear regression model to predict the full map of the future cycles. It assumes that the sensed data varies linearly over time.
  \item \textbf{WNN}, which combines wavelet transform and neural network. WNN is good at extracting periodic features of time series for data prediction.
  \item \textbf{NAR}, which uses a nonlinear autoregressive neural network to predict the near future. NAR considers the nonlinear temporal correlations within data.
\end{itemize}

\subsection{Completion on Extremely Sparse Data (RQ1)}

For most sparse data completion tasks, the sensing rate typically ranges from 20\% to 80\%, indicating that we can utilize abundant spatiotemporal information. However, in fine-grained and time-continuous completion, we have the matrices with sense ratio of $1/\text{n-columns}$ which traditional methods may have difficulty handling. To show the effectiveness of RNN-DMF on extremely sparse data, we compare it with other existing completion methods on multiple datasets. Note that the sense ratio is not limited to $1/\text{n-columns}$ to show the generalization performance of RNN-DMF.

As problem setting, each column of the data matrix represents a submission, and each row represents a subarea to sense. To build sparse datasets, we randomly mask the complete matrix and leave $1-5$ data points unmasked in each column to represent the sensed data. This process is entirely random, as cell selection strategy is not our focus. For fair comparison, we use a small amount of complete data to train dense-supervised methods to ensure that models function properly.

\begin{figure*}[htbp]
\centering
\begin{minipage}[t]{0.24\textwidth}
\centering
\includegraphics[width=4.7cm]{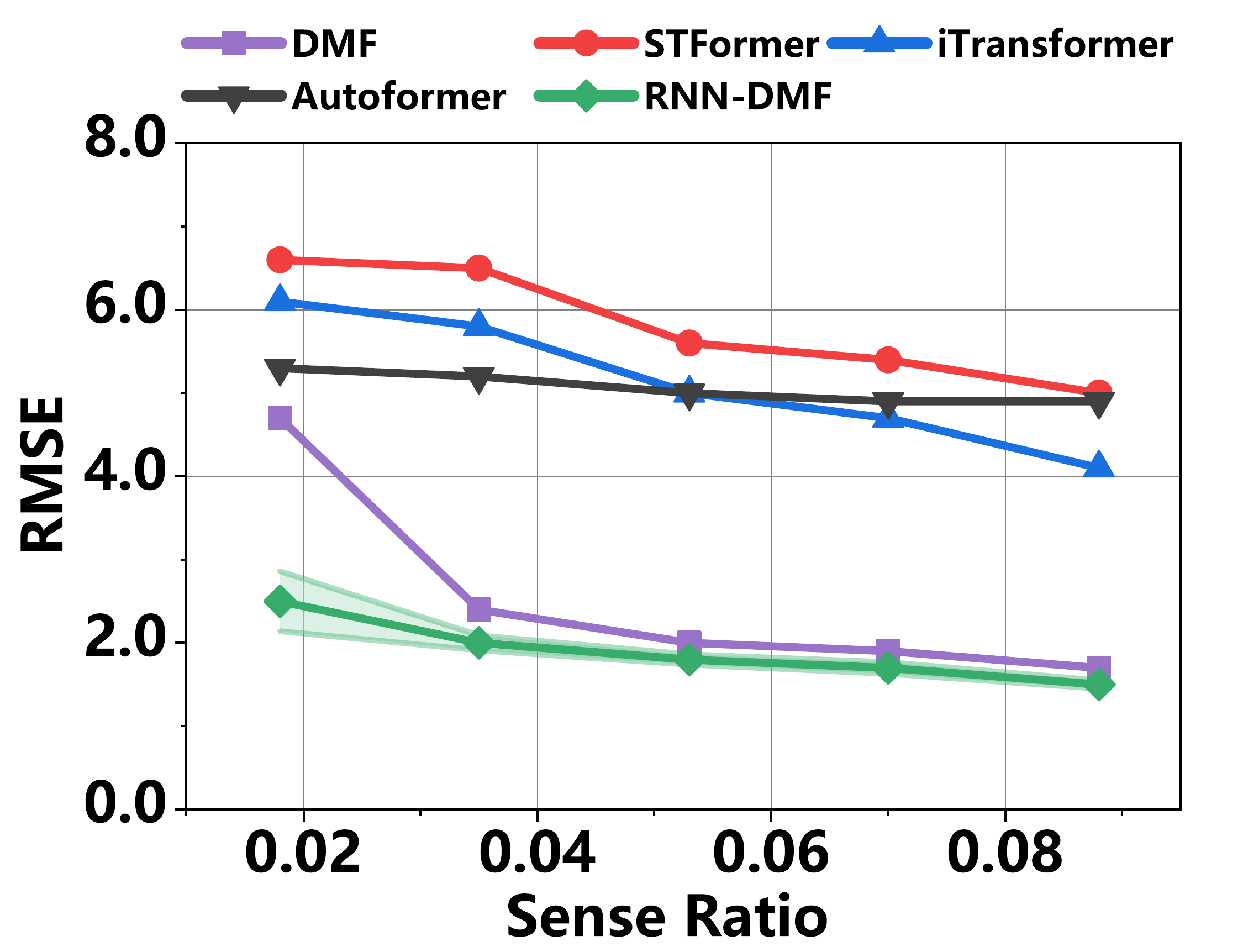}
\caption*{(a) U-AIR}
\end{minipage}
\begin{minipage}[t]{0.24\textwidth}
\centering
\includegraphics[width=4.7cm]{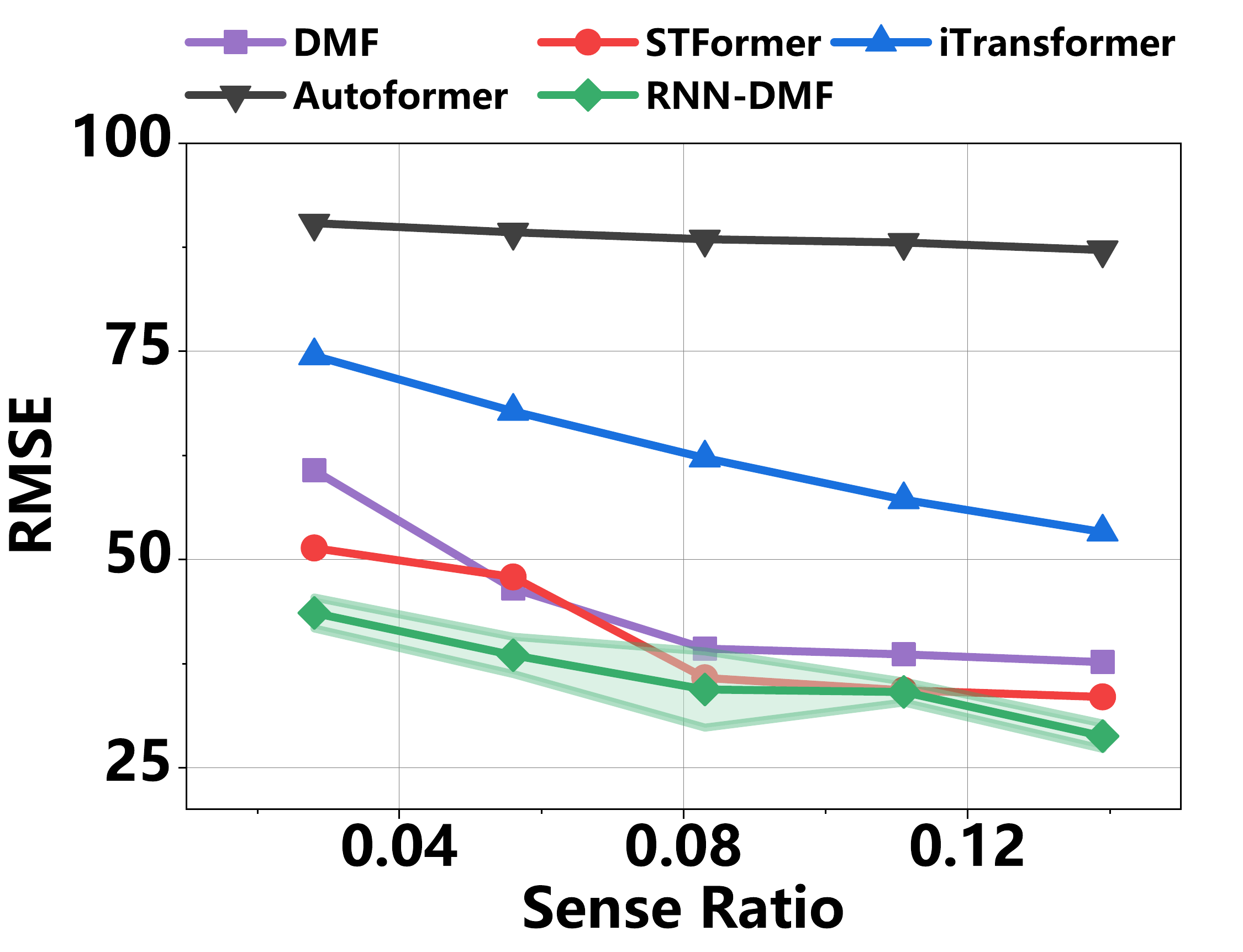}
\caption*{(b) Sensor-Scope}
\end{minipage}
\begin{minipage}[t]{0.24\textwidth}
\centering
\includegraphics[width=4.7cm]{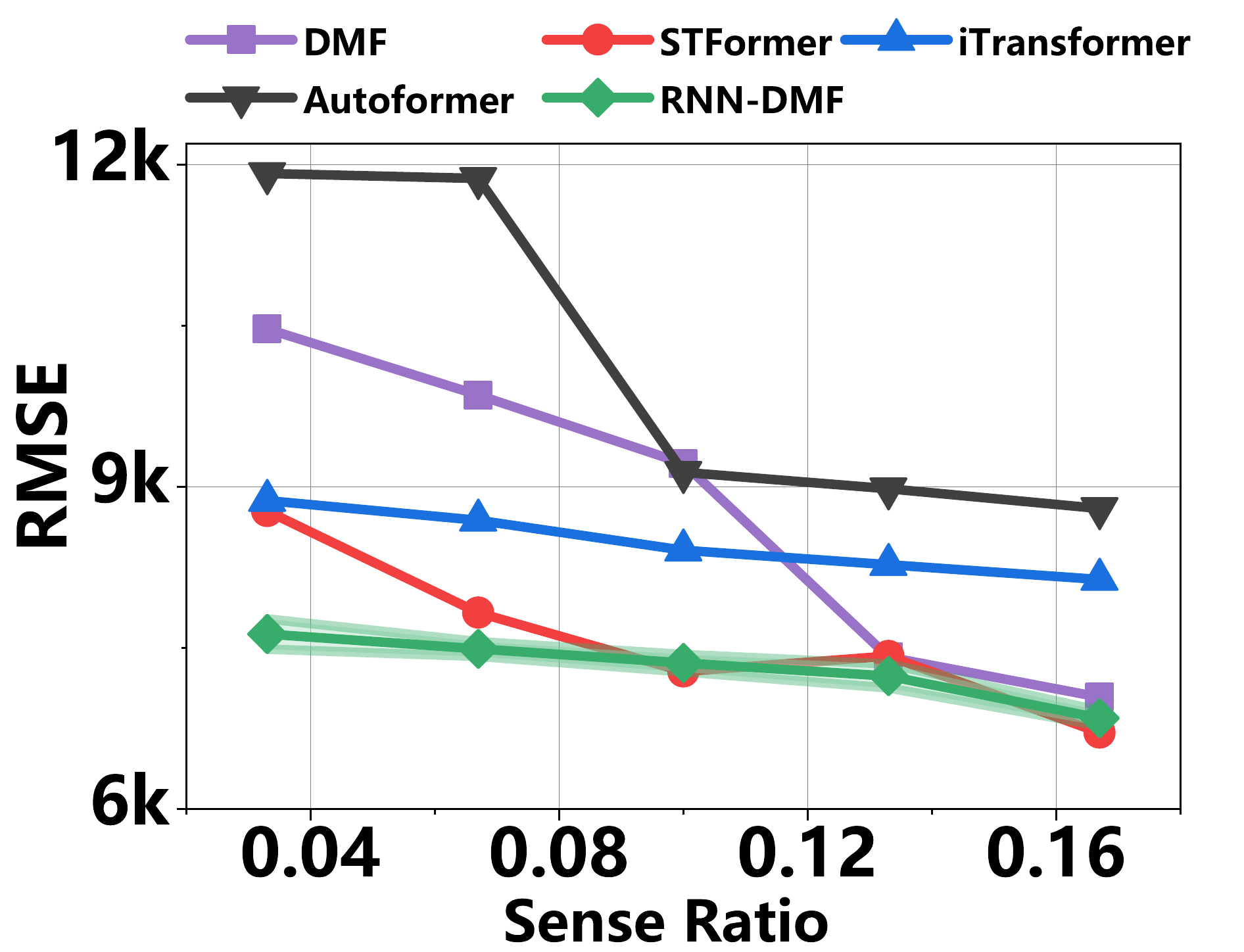}
\caption*{(c) TaxiSpeed}
\end{minipage}
\begin{minipage}[t]{0.24\textwidth}
\centering
\includegraphics[width=4.7cm]{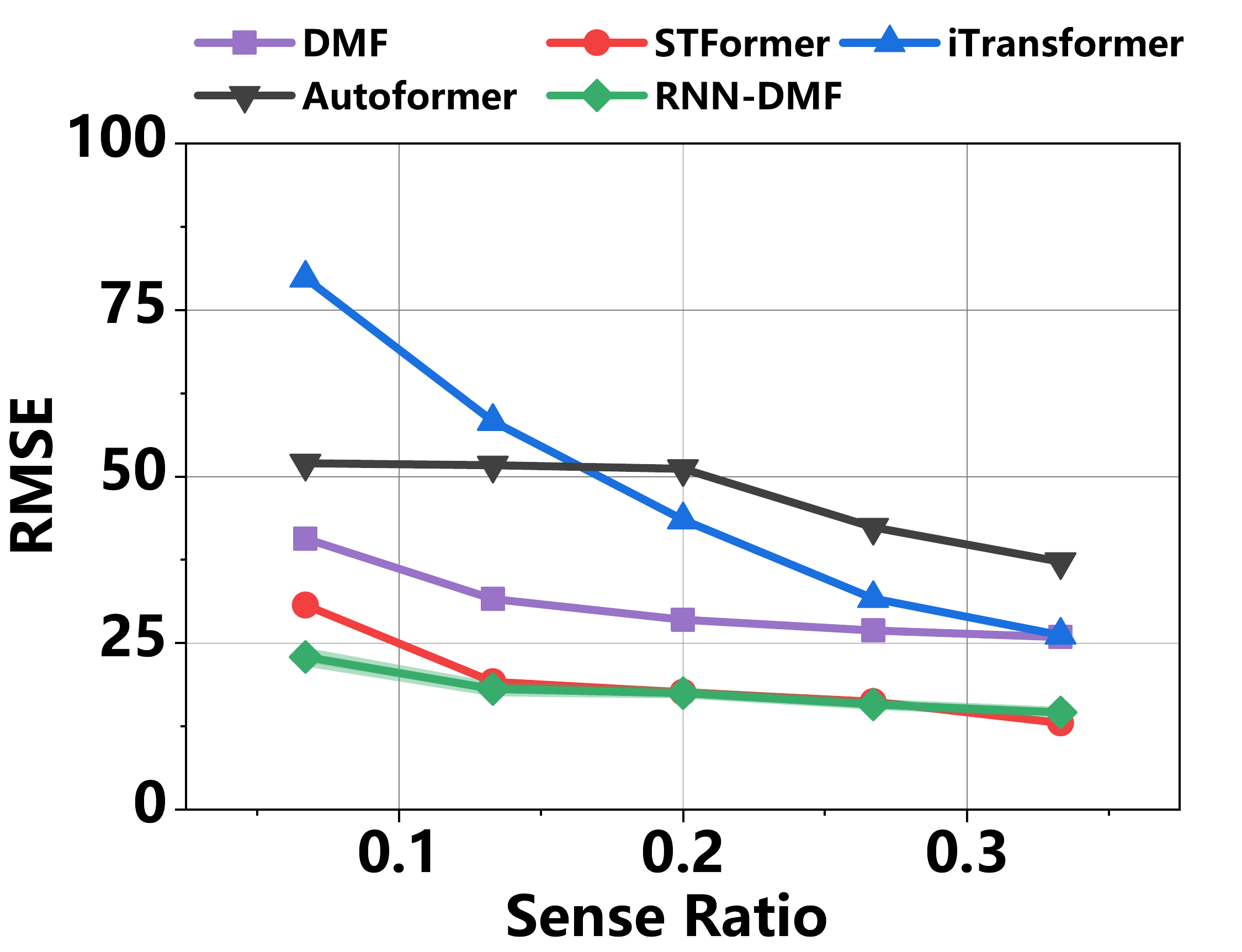}
\caption*{(d) Highways England}
\end{minipage}
\caption{Partial RMSE results of different data sparsity under different datasets.}
\label{fig_exp1}
\end{figure*}

The results in Table \ref{table_exp1} and Fig. \ref{fig_exp1} illustrate that RNN-DMF significantly outperforms existing works. The overall trend suggests that with the increase of sense ratio, the completion accuracy of all testing methods rises. This aligns with intuition as sensed data provides information for missing data inference. From Table \ref{table_exp1}, we can clearly see that in scenarios where data matrix is extremely sparse, our method surpasses most existing methods. The green error bars in Fig. \ref{fig_exp1} also highlight the robustness of our method. This experiment forms the basis for our work as both fine-grained completion and time-completion models deal with scenarios where spatial information is lacking and the matrix is extremely sparse. 

\begin{table*}[htbp]
  \centering
  \caption{Full RMSE results under different sparsity on four datasets. }
  \renewcommand{\arraystretch}{1.2} 
  \setlength{\tabcolsep}{8pt}
    \begin{tabular}{c|c|cccccc|cc}
    \hline
    \multirow{2}{*}{Dataset} & \multirow{2}{*}{Sensed Ratio} & \multicolumn{6}{c|}{Sparse-supervised method} & \multicolumn{2}{c}{Dense-supervised method} \\
    \cline{3-10}          &       & GP    & knn-s & MC    & DMF   & RNN-DMF & STformer & iTransformer & Autoformer \\
    \hline
    \multirow{5}[2]{*}{U-AIR} & 1/57 & 4.3  & 4.2  & 3.9   & 4.7   & \textbf{2.5} & 6.6   & 6.1   & 5.3 \\
          & 2/57 & 4.1  & 4.1  & 2.5   & 2.4   & \textbf{2.0} & 6.5   & 5.8   & 5.2 \\
          & 3/57 & 3.6 & 3.2  & 2.5   & 2.0     & \textbf{1.8} & 5.6   & 5.0     & 5.0 \\
          & 4/57  & 3.3  & 3.3  & 1.7   & 1.9   & \textbf{1.7} & 5.4   & 4.7   & 4.9 \\
          & 5/57 & 3.2  & 3.2  & 1.6   & 1.7   & \textbf{1.5} & 5.0     & 4.1   & 4.9 \\
    \hline
    \multirow{5}[2]{*}{Sensor-Scope} & 1/36 & 53.0   & 55.8  & 56.5  & 60.7  & \textbf{43.6} & 51.4  & 74.5  & 90.4 \\
          & 2/36 & 54.2  & 44.6  & 44.4  & 46.5  & \textbf{38.5} & 47.9  & 67.8  & 89.3 \\
          & 3/36 & 53.0    & 41.6  & 38.9  & 39.3  & \textbf{34.4} & 35.8  & 62.2  & 88.5 \\
          & 4/36 & 52.5  & 39.4  & 36.9  & 38.6  & \textbf{34.1} & 34.3  & 57.2  & 88.1 \\
          & 5/36 & 50.6  & 58.2  & 34.1  & 37.7  & \textbf{28.8} & 33.5  & 53.3  & 87.2 \\
    \hline
    \multirow{5}[2]{*}{TaxiSpeed} & 1/30 & 39028.1 & 38252.8 & 31450.2 & 10474.5 & \textbf{7632.0} & 8778.2  & 8876.7  & 11921.1 \\
          & 2/30 & 39990.3 & 32182.1 & 28295.3 & 9859.5  & \textbf{7493.1} & 7831.3  & 8691.6  & 11876.0 \\
          & 3/30   & 38637.5 & 29491.2 & 24141.7 & 9220.7  & 7361.5  & \textbf{7283.9} & 8414.7  & 9139.1 \\
          & 4/30 & 36617.3 & 28337.3 & 22709.9 & 7417.5  & \textbf{7241.1} & 7428.0  & 8277.5  & 8984.9 \\
          & 5/30 & 35991.2 & 26787.2 & 21344.1 & 7012.0  & 6849.1  & \textbf{6713.5} & 8139.8  & 8803.8 \\
    \hline
    \multirow{5}[2]{*}{Hishways England} & 1/15   & 66.5  & 70.5  & 66.1  & 40.7  & \textbf{22.9} & 30.7  & 79.8  & 52.0 \\
          & 2/15  & 56.2  & 60.3  & 49.6  & 31.6  & \textbf{18.1} & 19.2  & 58.3  & 51.7 \\
          & 3/15    & 50.7  & 53.8  & 41.6  & 28.5  & \textbf{17.5} & 17.6  & 43.5  & 51.2 \\
          & 4/15  & 45.9  & 49.9  & 36.9  & 26.9  & \textbf{15.8} & 16.2  & 31.7  & 42.4 \\
          & 5/15  & 42.5  & 52.4  & 29.8  & 25.9  & 14.6  & \textbf{13.0} & 26.2  & 37.2 \\
    \hline
    \end{tabular}%
  \label{table_exp1}%
\end{table*}

\subsection{Ablation Study on Time Gates (RQ2)}
Based on RNN-DMF, we aim to further demonstrate the effectiveness of time gates which are designed to capture information within unequal intervals between submissions. We design an ablation experiment between TIME-DMF, which incorporates time gates, and RNN-DMF, which does not incorporate time gates, on multiple datasets. 

As most publicly available datasets consist of sensor data with equal time intervals, we artificially create datasets through random masking and deletion. In specific, for the complete data matrix, we randomly remove some columns from the matrix so that the left columns are of random intervals. For the columns left, we do random masking again like that in the last experiment. This preprocessing results in extremely sparse matrices where the intervals between columns vary in length. Because of the deletion, the datasets will be much smaller. So we need datasets of enough columns to make sure the experiments can have stable results.  As Highways England is a dataset that collects road data per 15 minutes for years, it provides an ample number of columns for use. In this experiment we employ this typical dataset to test the effectiveness of time gates.

\begin{figure}[tbp]
\centering
\begin{minipage}[t]{0.155\textwidth}
\centering
\includegraphics[width=3.4cm]{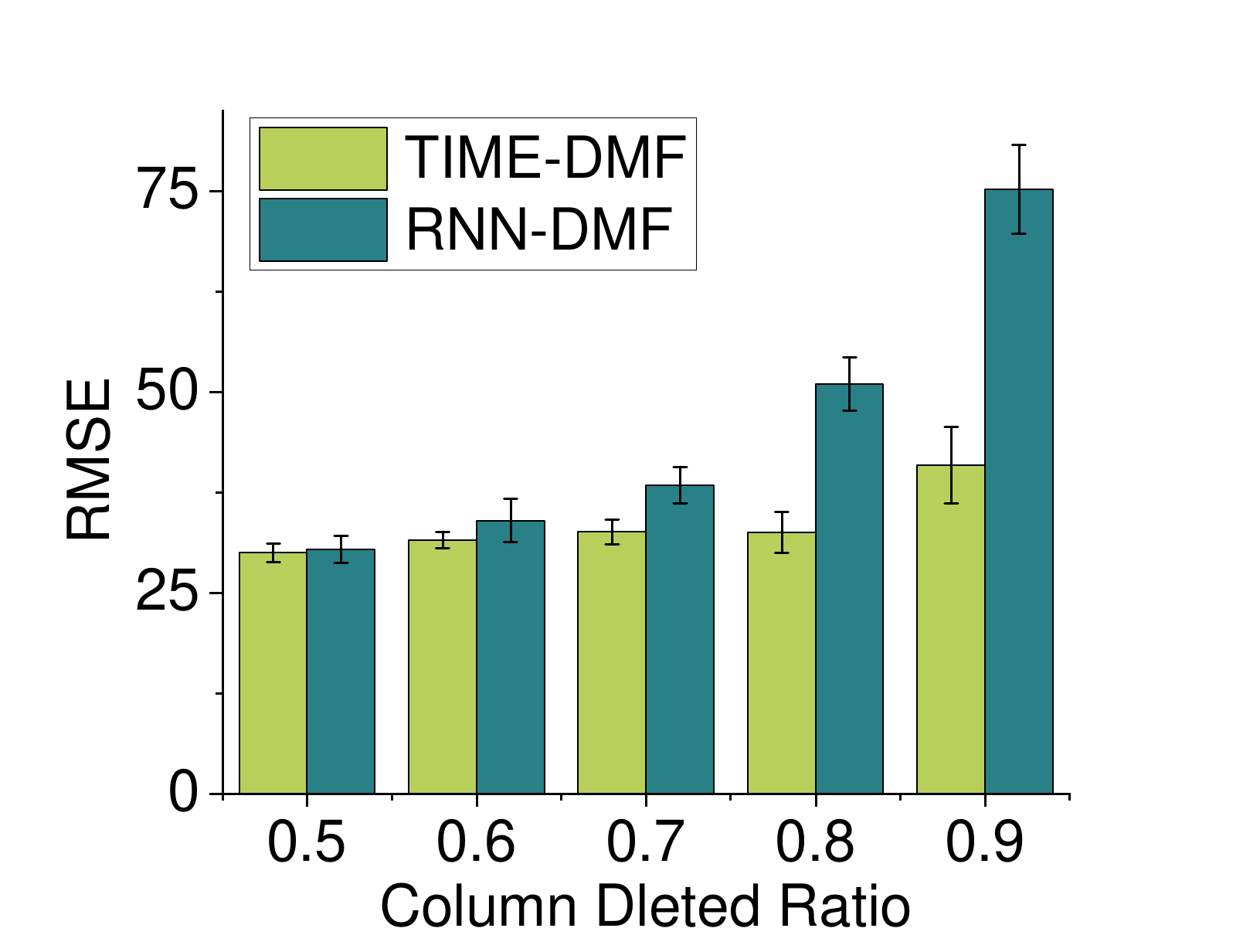}
\caption*{(a) 2021-1month}
\end{minipage}
\begin{minipage}[t]{0.155\textwidth}
\centering
\includegraphics[width=3.4cm]{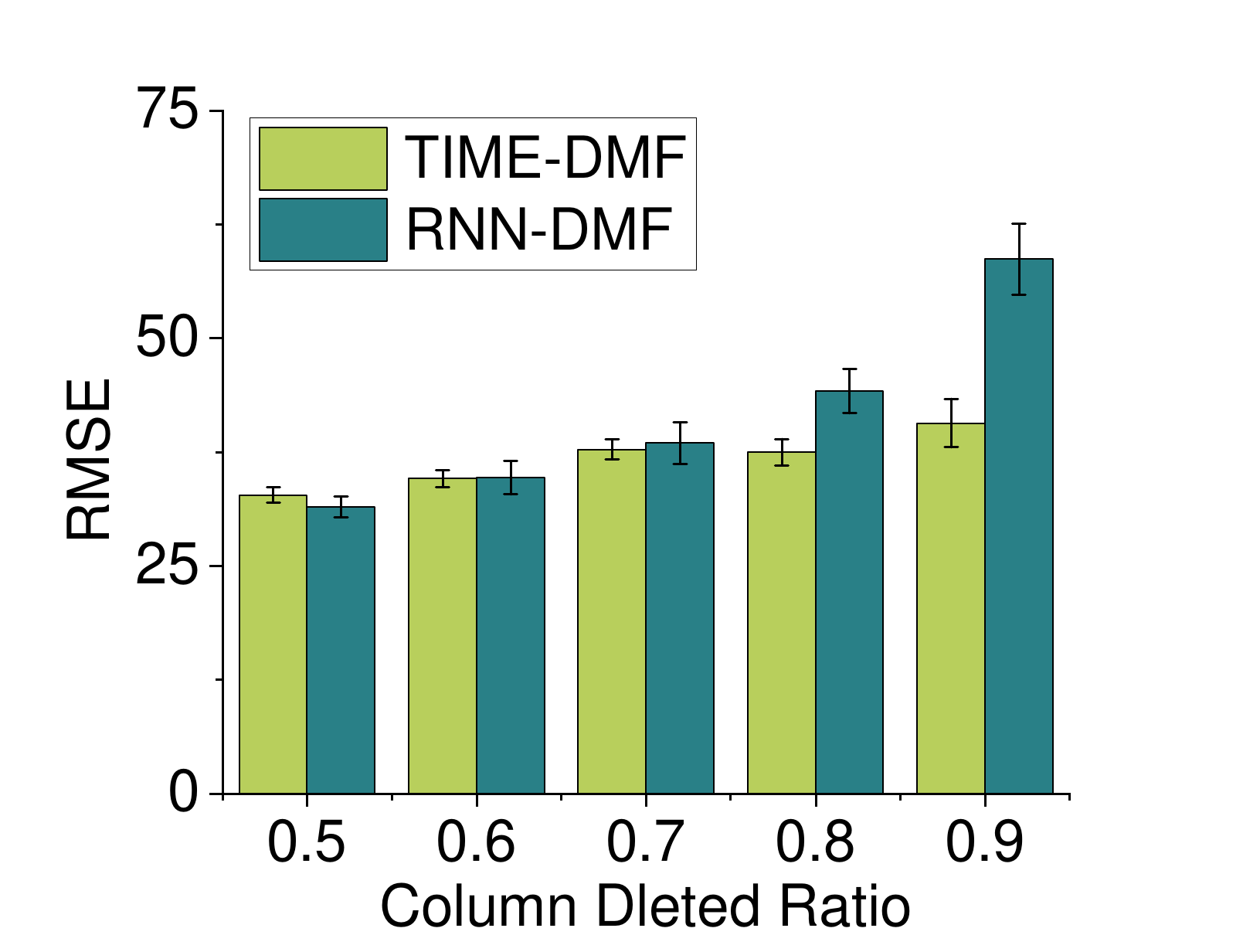}
\caption*{(b) 2021-2months}
\end{minipage}
\begin{minipage}[t]{0.155\textwidth}
\centering
\includegraphics[width=3.4cm]{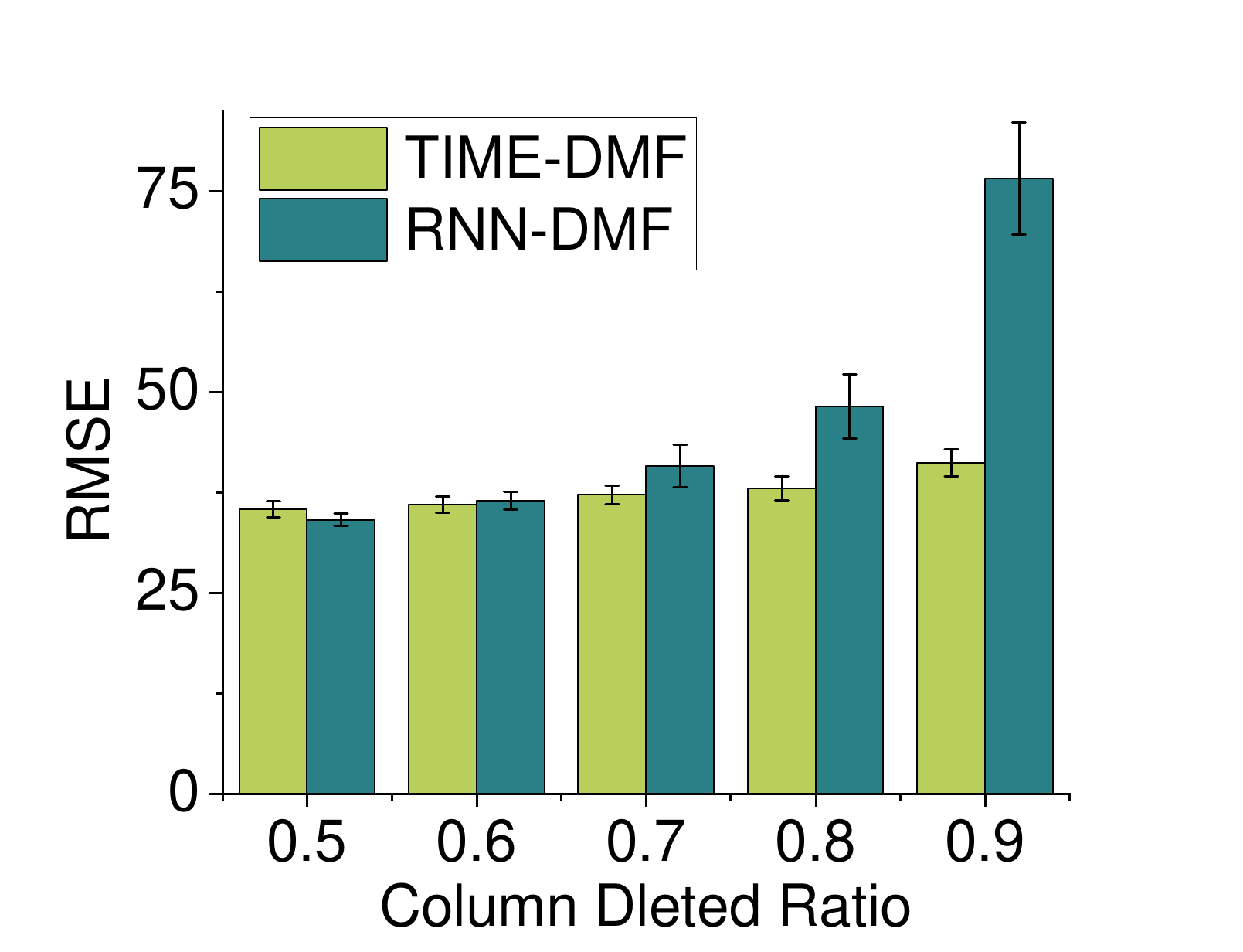}
\caption*{(c) 2021-3months}
\end{minipage}
\begin{minipage}[t]{0.155\textwidth}
\centering
\includegraphics[width=3.4cm]{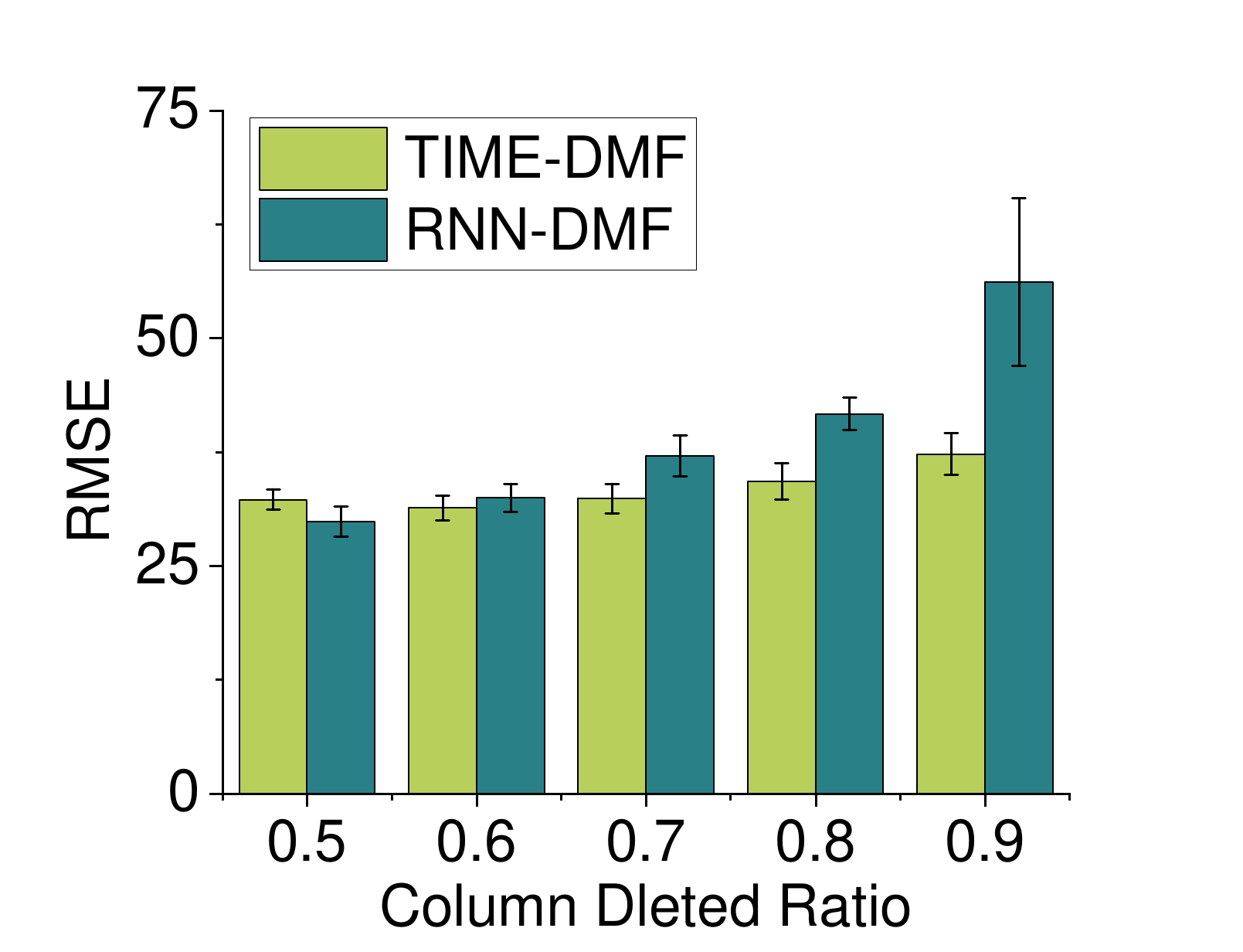}
\caption*{(d) 2022-2weeks}
\end{minipage}
\begin{minipage}[t]{0.155\textwidth}
\centering
\includegraphics[width=3.4cm]{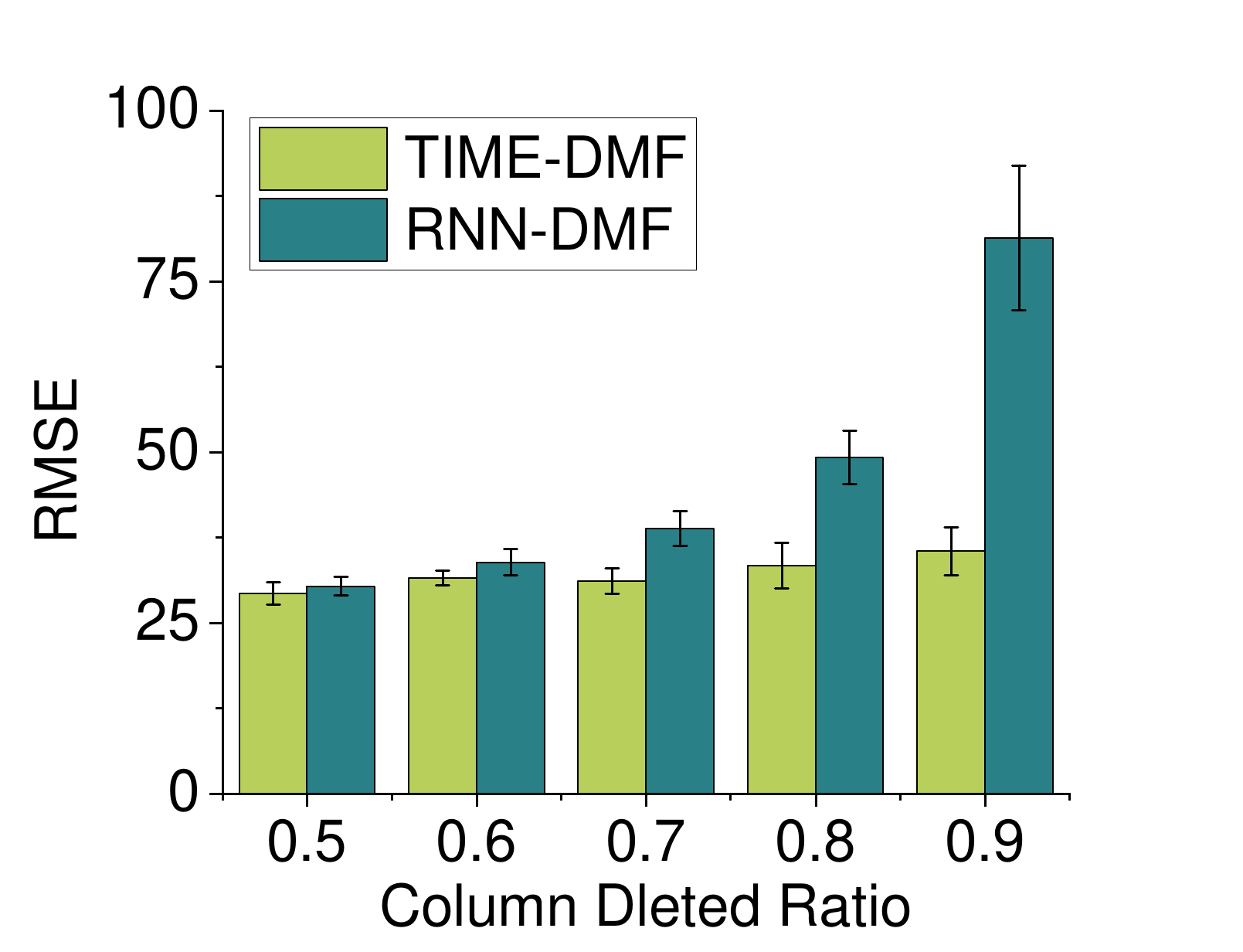}
\caption*{(e) 2022-1month}
\end{minipage}
\begin{minipage}[t]{0.155\textwidth}
\centering
\includegraphics[width=3.4cm]{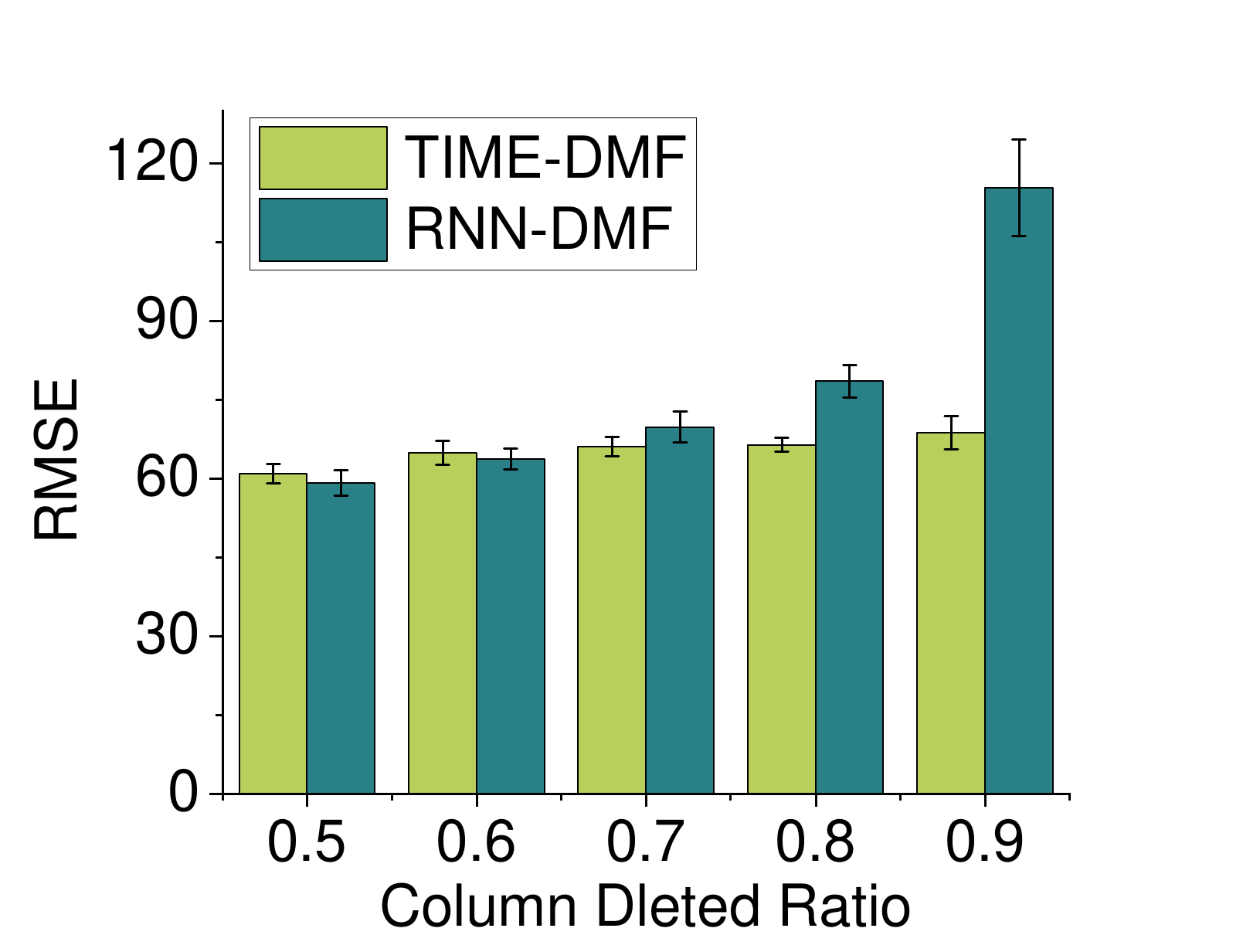}
\caption*{(f) 2022-2months}
\end{minipage}
\caption{Completion results of two models with the only
difference of time gates.}
\label{fig_exp2}
\end{figure}

The experimental results are shown in Table \ref{table_exp2} and Fig. \ref{fig_exp2}. The x-axis represents the proportion of columns deleted from the original dataset. When the deletion proportion grows bigger, the interval between columns will be more uneven and there will be more temporal information within time intervals for time gates to capture. In all datasets, as the unevenness of the data increases, the inference performance of both RNN-DMF and TIME-DMF decreases. This is because when the data is uneven, the similarity between adjacent submissions decreases and it's hard for encoders to capture direct correlations between submissions. This trend is particularly evident when the deletion ratio is high. However, compared with RNN-DMF, TIME-DMF performs much more stably when the deletion rate and data unevenness increases. This is because RNN-DMF can only statically share parameters between time steps without considering the interval length. While in TIME-DMF, the global memory and local memory can cooperate to dynamically control the flow of temporal information. This propagation pattern better utilizes the information within intervals of different lengths.
If we consider RNN-DMF as the baseline, we observe that TIME-DMF performs better when other conditions remain unchanged. This directly proves the effectiveness of time gates in capturing information within unequal intervals.

\begin{table*}[t]
\centering
\caption{RMSE results of the ablation study of time gates.}
\renewcommand\arraystretch{1.2}
\begin{minipage}{0.49\textwidth}
\centering
\scalebox{0.97}{
\begin{tabular}{ccccccc}
\toprule
\multirow{2}{*}{Dataset} & \multicolumn{1}{c}{\multirow{2}{*}{Method}} & \multicolumn{5}{c}{Slices Deleted Ratio} \\ \cline{3-7} 
 & \multicolumn{1}{c}{} & \multicolumn{1}{c}{0.5} & \multicolumn{1}{c}{0.6} & \multicolumn{1}{c}{0.7} & \multicolumn{1}{c}{0.8} & 0.9 \\ \hline
\multirow{2}{*}{1month} & RNN-DMF & \multicolumn{1}{c}{30.4} & \multicolumn{1}{c}{34.0} & \multicolumn{1}{c}{38.4} & \multicolumn{1}{c}{51.0} & 75.2 \\ \cline{2-2} 
 & TIME-DMF & \multicolumn{1}{c}{\textbf{29.2}} & \multicolumn{1}{c}{\textbf{29.0}} & \multicolumn{1}{c}{\textbf{31.4}} & \multicolumn{1}{c}{\textbf{32.5}} & \textbf{42.8} \\ \hline
\multirow{2}{*}{2months} & RNN-DMF & \multicolumn{1}{c}{\textbf{31.5}} & \multicolumn{1}{c}{34.7} & \multicolumn{1}{c}{38.5} & \multicolumn{1}{c}{44.2} & 58.7 \\ \cline{2-2} 
 & TIME-DMF & \multicolumn{1}{c}{32.8} & \multicolumn{1}{c}{\textbf{34.6}} & \multicolumn{1}{c}{\textbf{37.8}} & \multicolumn{1}{c}{\textbf{37.5}} & \textbf{40.7} \\ \hline
\multirow{2}{*}{3months} & RNN-DMF & \multicolumn{1}{c}{\textbf{34.1}} & \multicolumn{1}{c}{36.5} & \multicolumn{1}{c}{40.8} & \multicolumn{1}{c}{48.2} & 76.6 \\ \cline{2-2} 
 & TIME-DMF & \multicolumn{1}{c}{35.4} & \multicolumn{1}{c}{\textbf{36.0}} & \multicolumn{1}{c}{\textbf{37.2}} & \multicolumn{1}{c}{\textbf{38.0}} & \textbf{41.2} \\ \bottomrule
\end{tabular}}
\caption*{\small Highways England for 2021}
\end{minipage}
\centering
\begin{minipage}{0.49\textwidth}
\scalebox{0.97}{
\begin{tabular}{ccccccc}
\toprule
\multirow{2}{*}{Dataset} & \multicolumn{1}{c}{\multirow{2}{*}{Method}} & \multicolumn{5}{c}{Slices Deleted Ratio} \\ \cline{3-7} 
 & \multicolumn{1}{c}{} & \multicolumn{1}{c}{0.5} & \multicolumn{1}{c}{0.6} & \multicolumn{1}{c}{0.7} & \multicolumn{1}{c}{0.8} & 0.9 \\ \hline
\multirow{2}{*}{2weeks} & RNN-DMF & \multicolumn{1}{c}{\textbf{29.9}} & \multicolumn{1}{c}{32.5} & \multicolumn{1}{c}{37.1} & \multicolumn{1}{c}{41.7} & 56.2 \\ \cline{2-2} 
 & TIME-DMF & \multicolumn{1}{c}{32.3} & \multicolumn{1}{c}{\textbf{31.4}} & \multicolumn{1}{c}{\textbf{32.4}} & \multicolumn{1}{c}{\textbf{34.3}} & \textbf{37.3} \\ \hline
\multirow{2}{*}{1month} & RNN-DMF & \multicolumn{1}{c}{30.4} & \multicolumn{1}{c}{33.9} & \multicolumn{1}{c}{38.8} & \multicolumn{1}{c}{49.2} & 81.4 \\ \cline{2-2} 
 & TIME-DMF & \multicolumn{1}{c}{\textbf{29.3}} & \multicolumn{1}{c}{\textbf{31.6}} & \multicolumn{1}{c}{\textbf{31.1}} & \multicolumn{1}{c}{\textbf{33.4}} & \textbf{35.5} \\ \hline
\multirow{2}{*}{2months} & RNN-DMF & \multicolumn{1}{c}{\textbf{59.2}} & \multicolumn{1}{c}{\textbf{3.7}} & \multicolumn{1}{c}{69.8} & \multicolumn{1}{c}{78.5} & 115.3 \\ \cline{2-2} 
 & TIME-DMF & \multicolumn{1}{c}{60.9} & \multicolumn{1}{c}{64.9} & \multicolumn{1}{c}{\textbf{66.1}} & \multicolumn{1}{c}{\textbf{66.4}} & \textbf{68.7} \\ \bottomrule
\end{tabular}}
\caption*{\small Highways England for 2022}
\end{minipage}
\label{table_exp2}
\end{table*}

\subsection{Demonstration of Generative Capability (RQ3)}

\begin{figure*}[tbp]
\centering
\begin{minipage}[t]{0.24\textwidth}
\centering
\includegraphics[width=4.7cm]{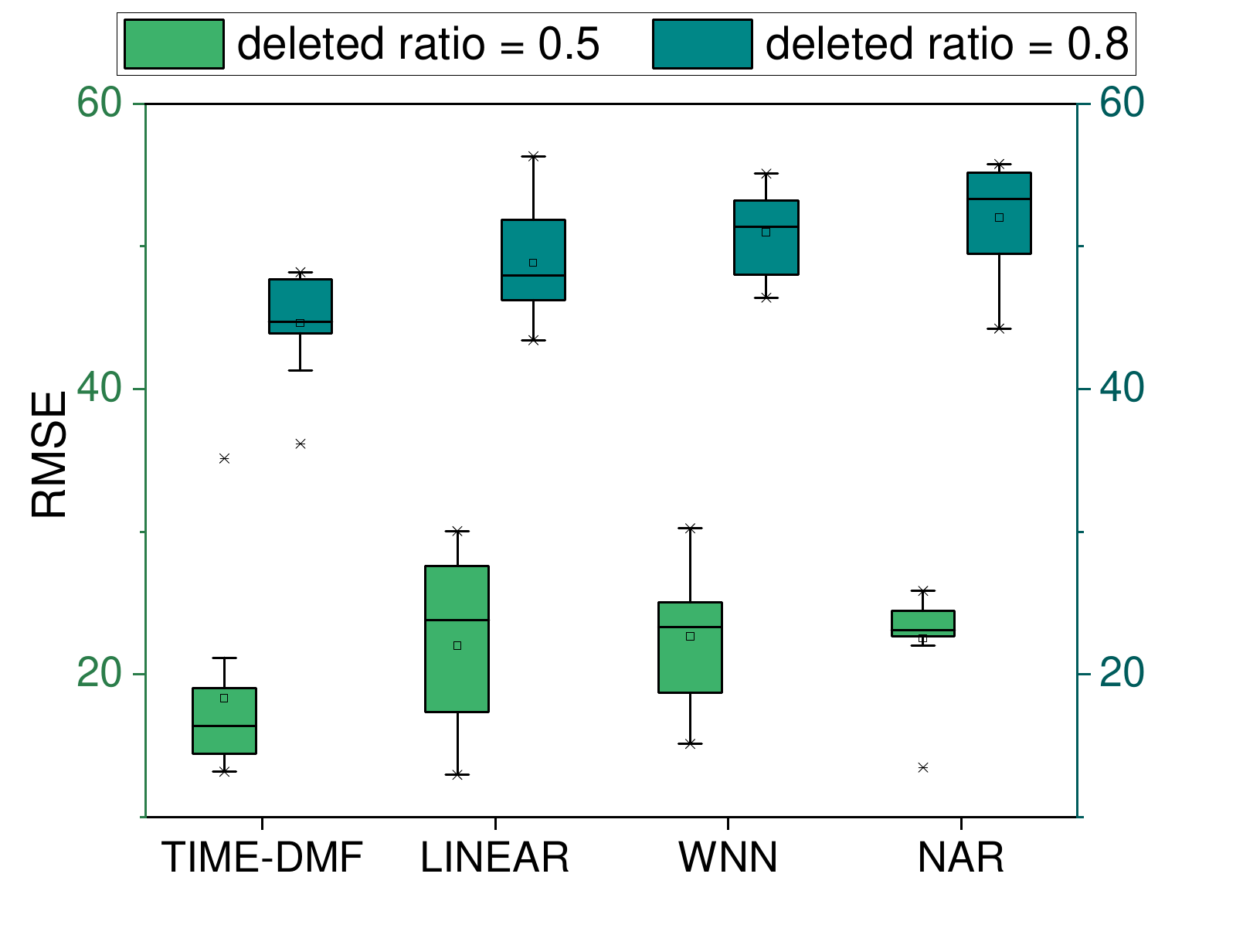}
\caption*{(a) U-AIR}
\end{minipage}
\begin{minipage}[t]{0.24\textwidth}
\centering
\includegraphics[width=4.7cm]{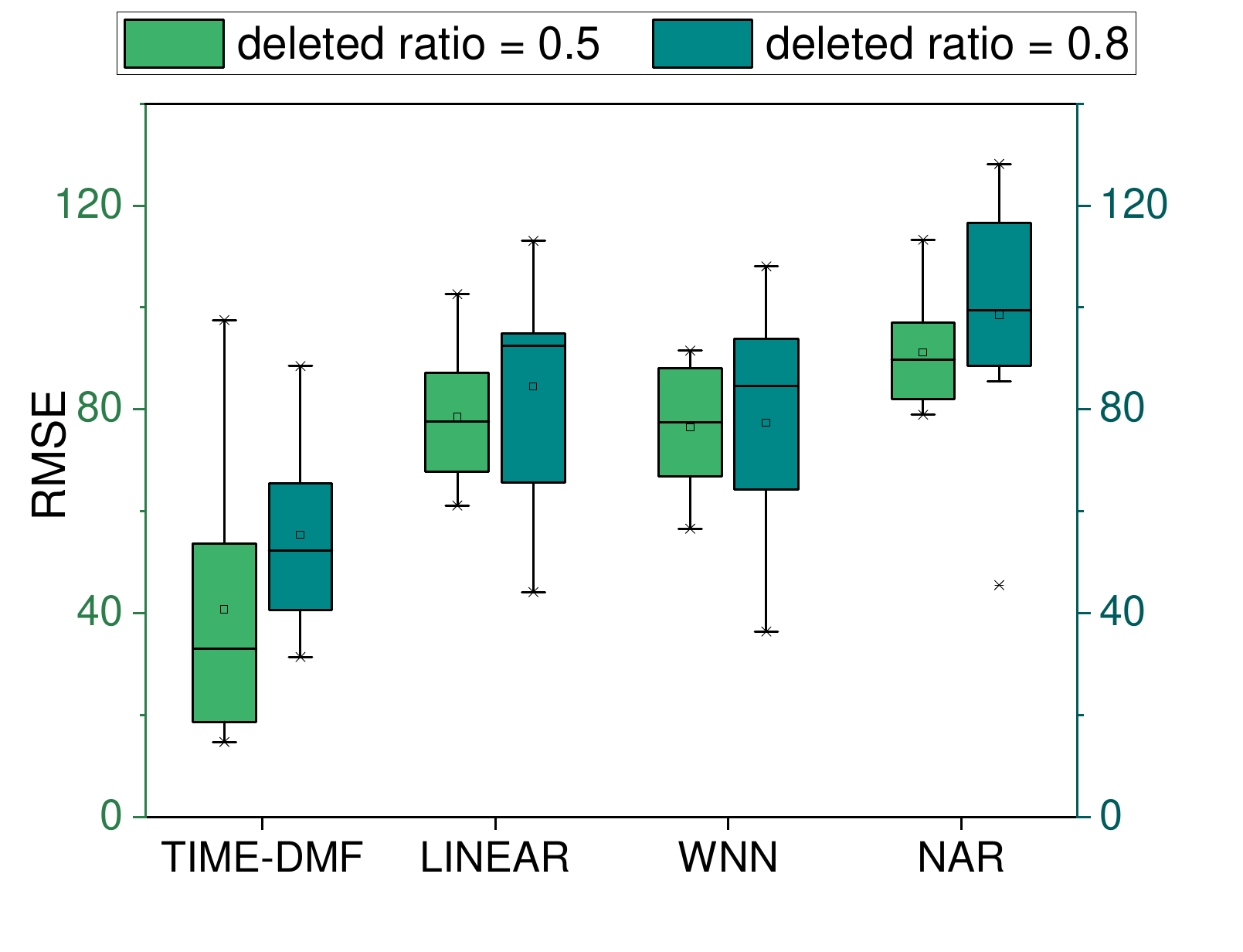}
\caption*{(b) Sensor-Scope}
\end{minipage}
\begin{minipage}[t]{0.24\textwidth}
\centering
\includegraphics[width=4.7cm]{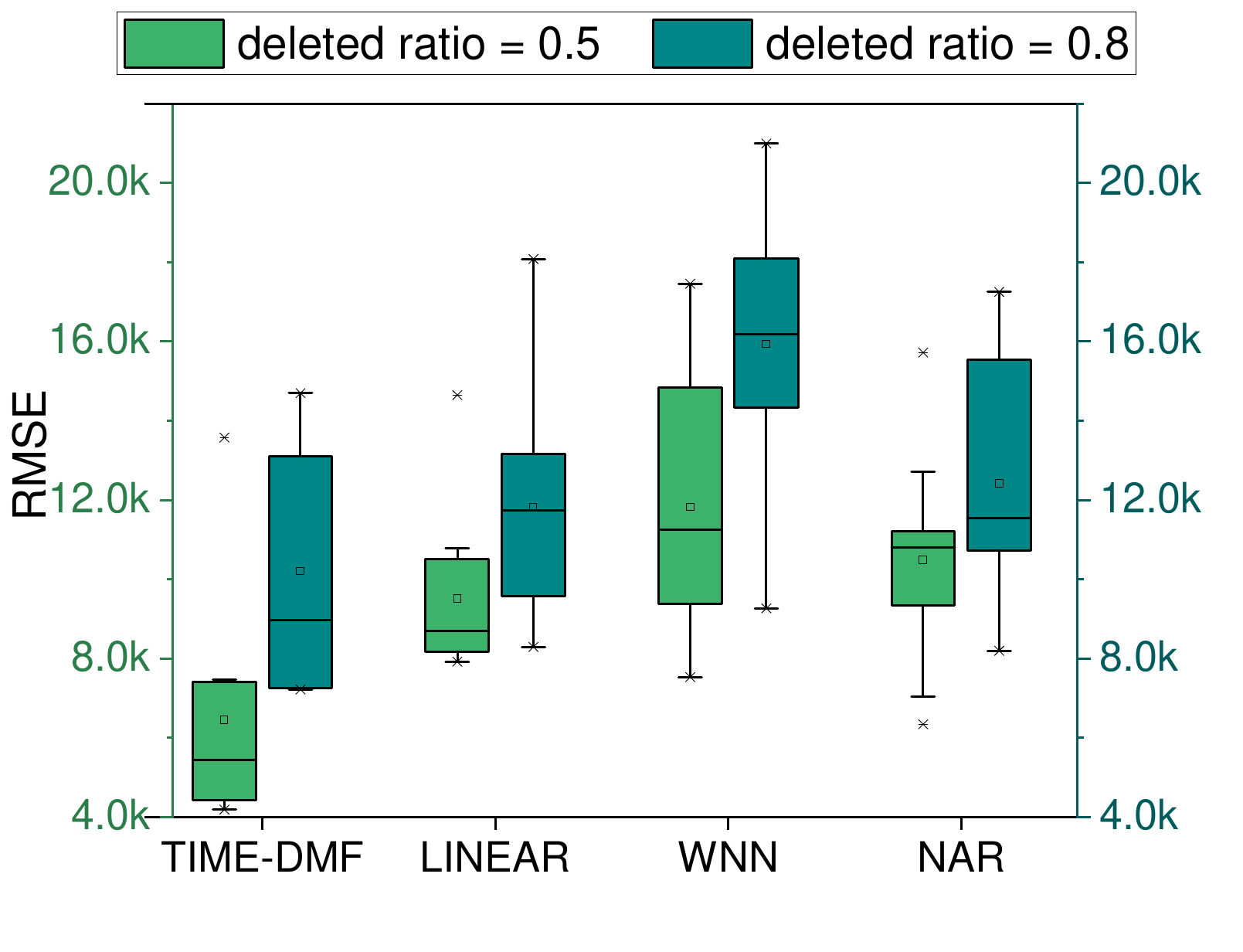}
\caption*{(c) TaxiSpeed}
\end{minipage}
\begin{minipage}[t]{0.24\textwidth}
\centering
\includegraphics[width=4.7cm]{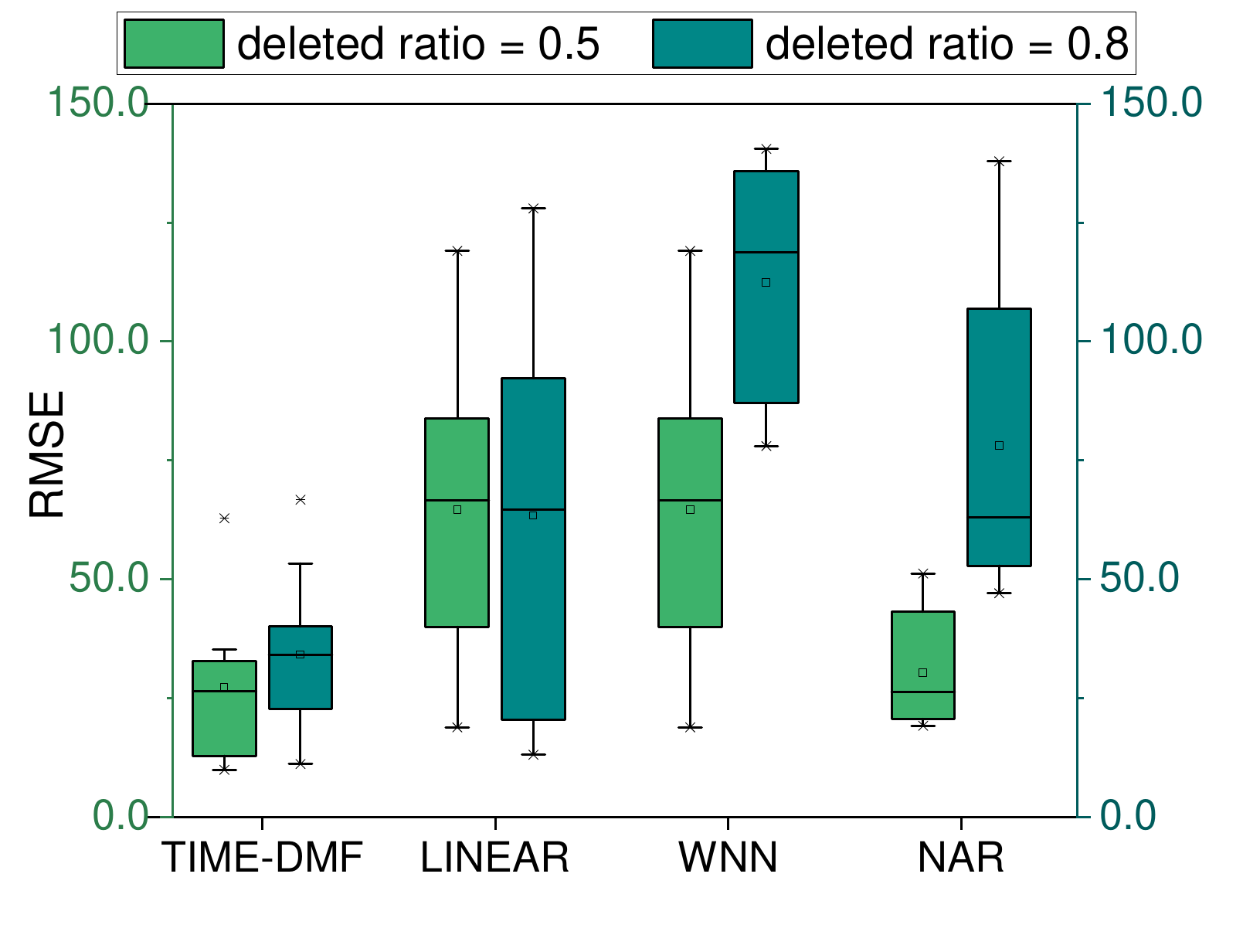}
\caption*{(d) Highways England}
\end{minipage}
\caption{RMSE of different data unevenness under different datasets.}
\label{fig_exp4}
\end{figure*}

\begin{figure*}
\centering

\begin{minipage}{\columnwidth}
\centering 

\begin{minipage}[t]{0.32\columnwidth}
\centering
\includegraphics[width=3.3cm]{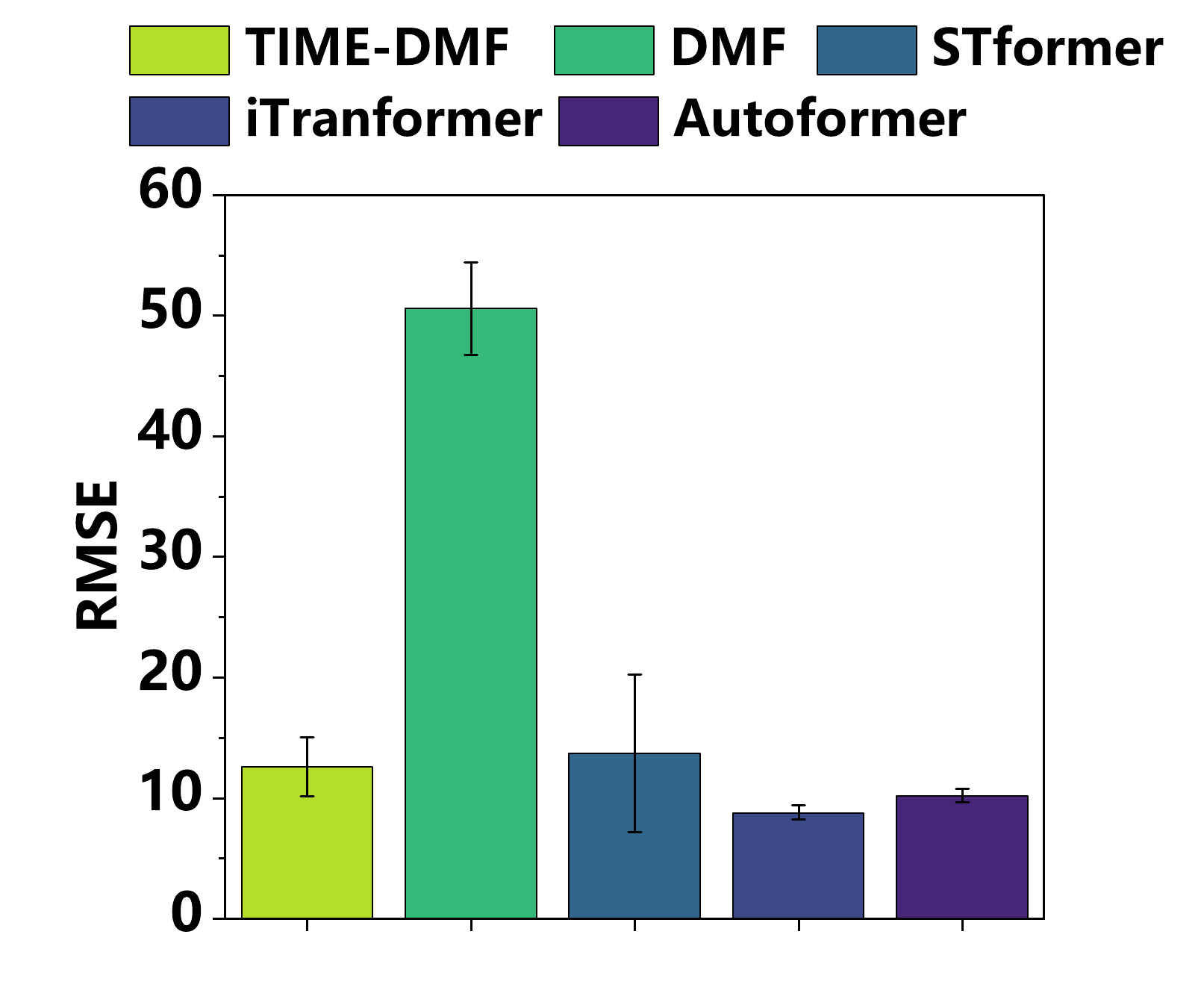}
\caption*{(a) U-AIR}
\end{minipage}
\begin{minipage}[t]{0.32\columnwidth}
\centering
\includegraphics[width=3.3cm]{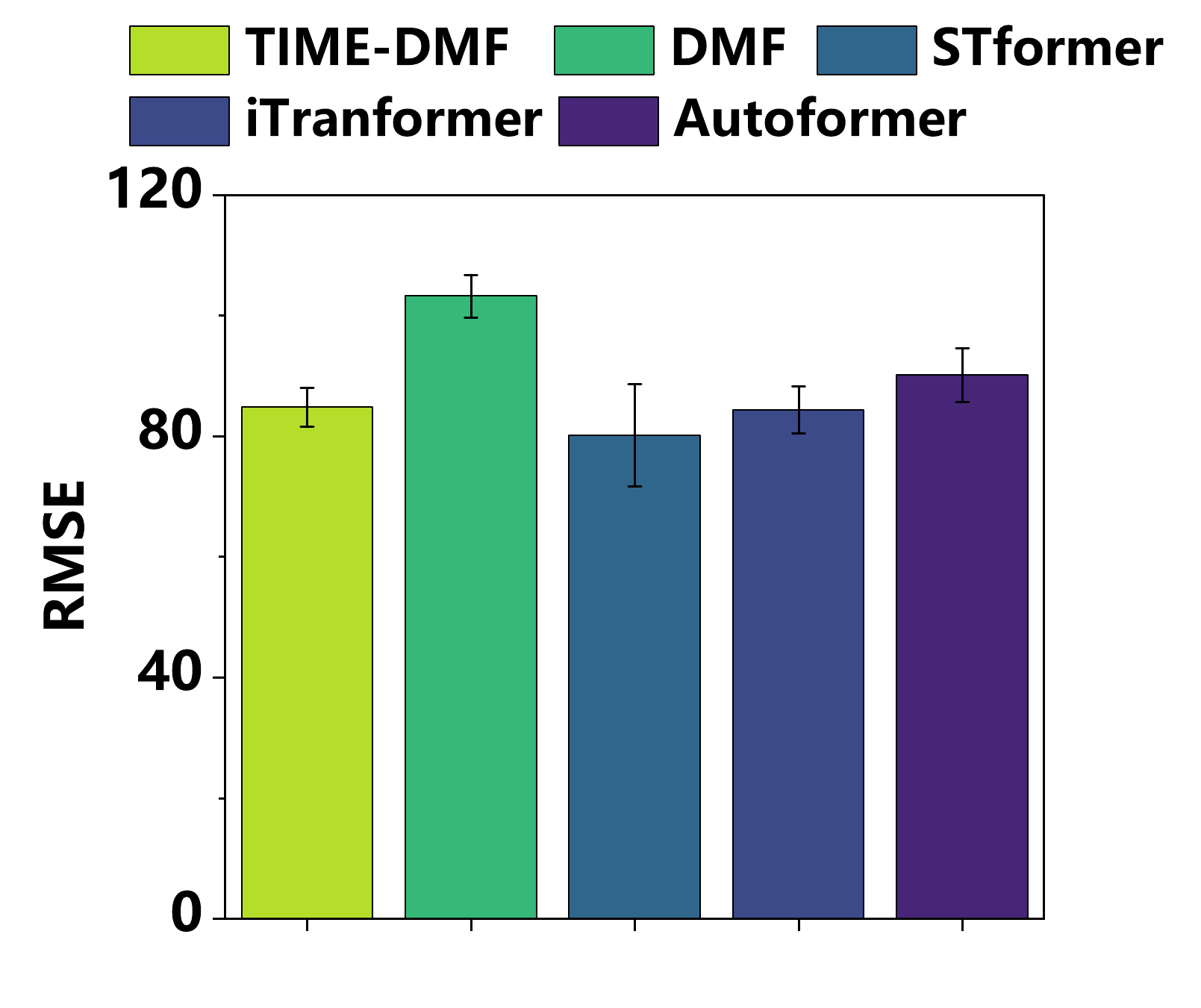}
\caption*{(b) Sensor-Scope}
\end{minipage}
\begin{minipage}[t]{0.32\columnwidth}
\centering
\includegraphics[width=3.3cm]{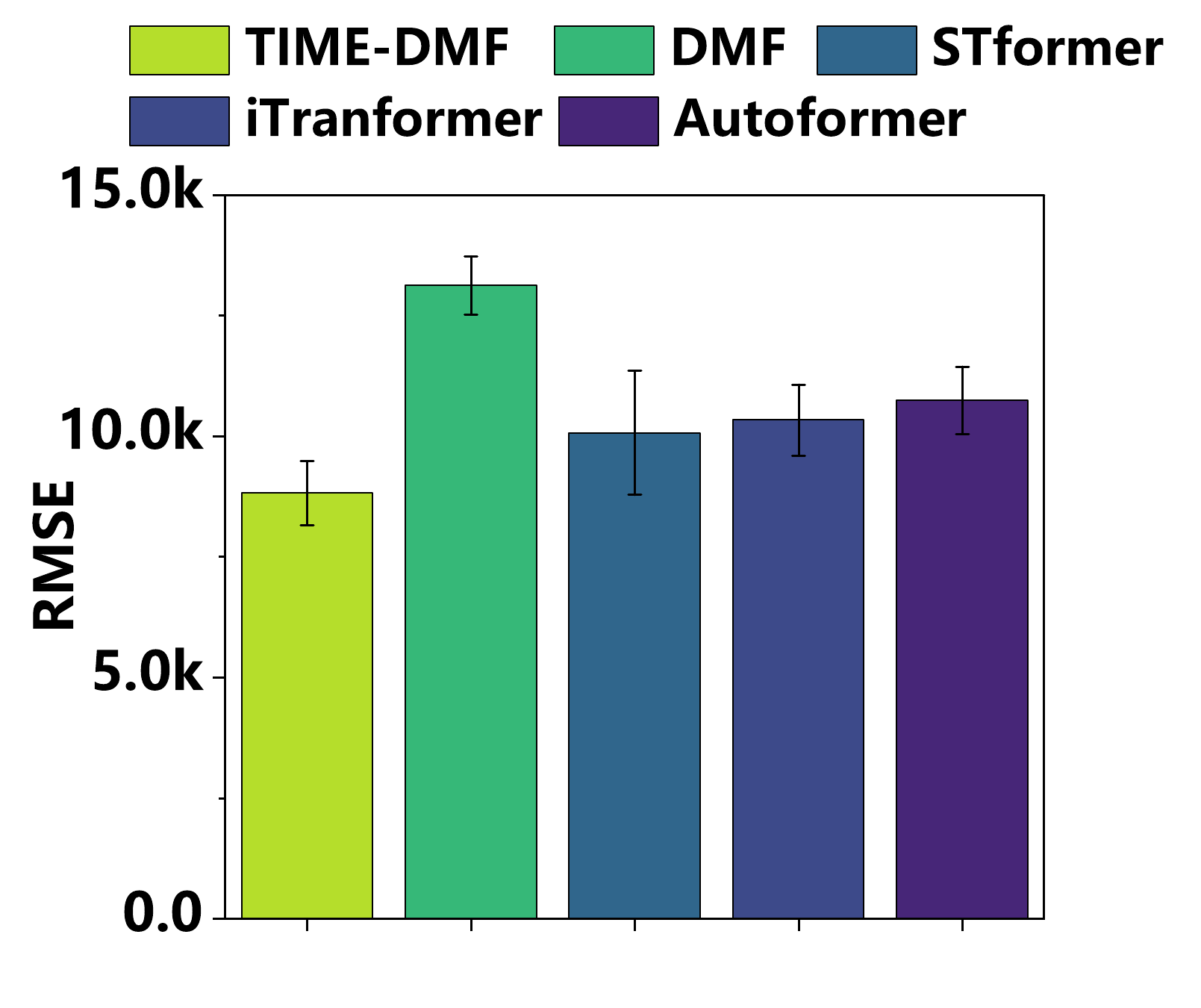}
\caption*{(c) TaxiSpeed}
\end{minipage}
\caption{RMSE for tiny datasets.}
\label{fig_exp3.1}
\end{minipage}
\begin{minipage}{\columnwidth}
\centering

\begin{minipage}[t]{.32\columnwidth}
\centering
\includegraphics[width=3.3cm]{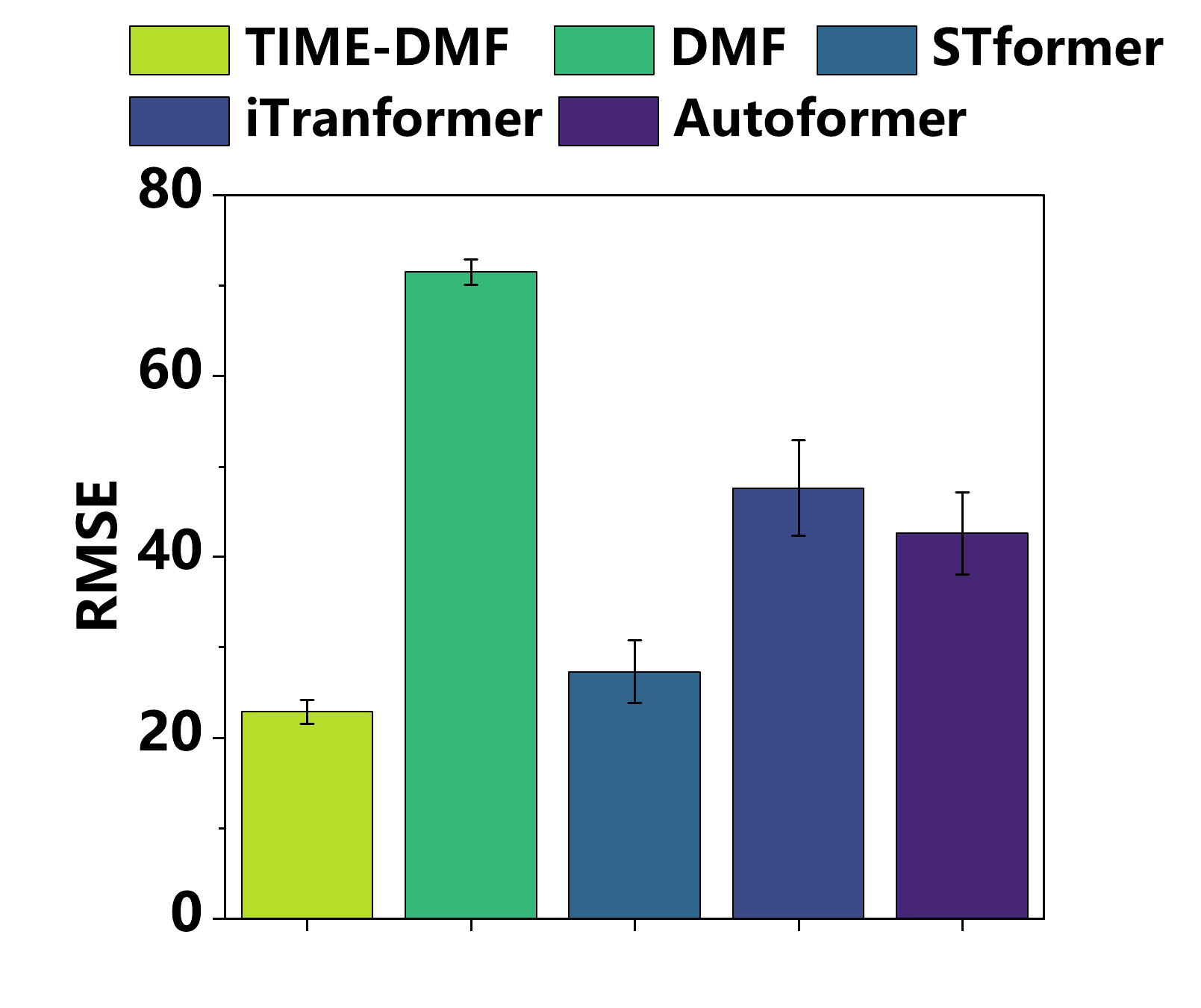}
\caption*{(a) HE-1month}
\end{minipage}
\begin{minipage}[t]{.32\columnwidth}
\centering
\includegraphics[width=3.3cm]{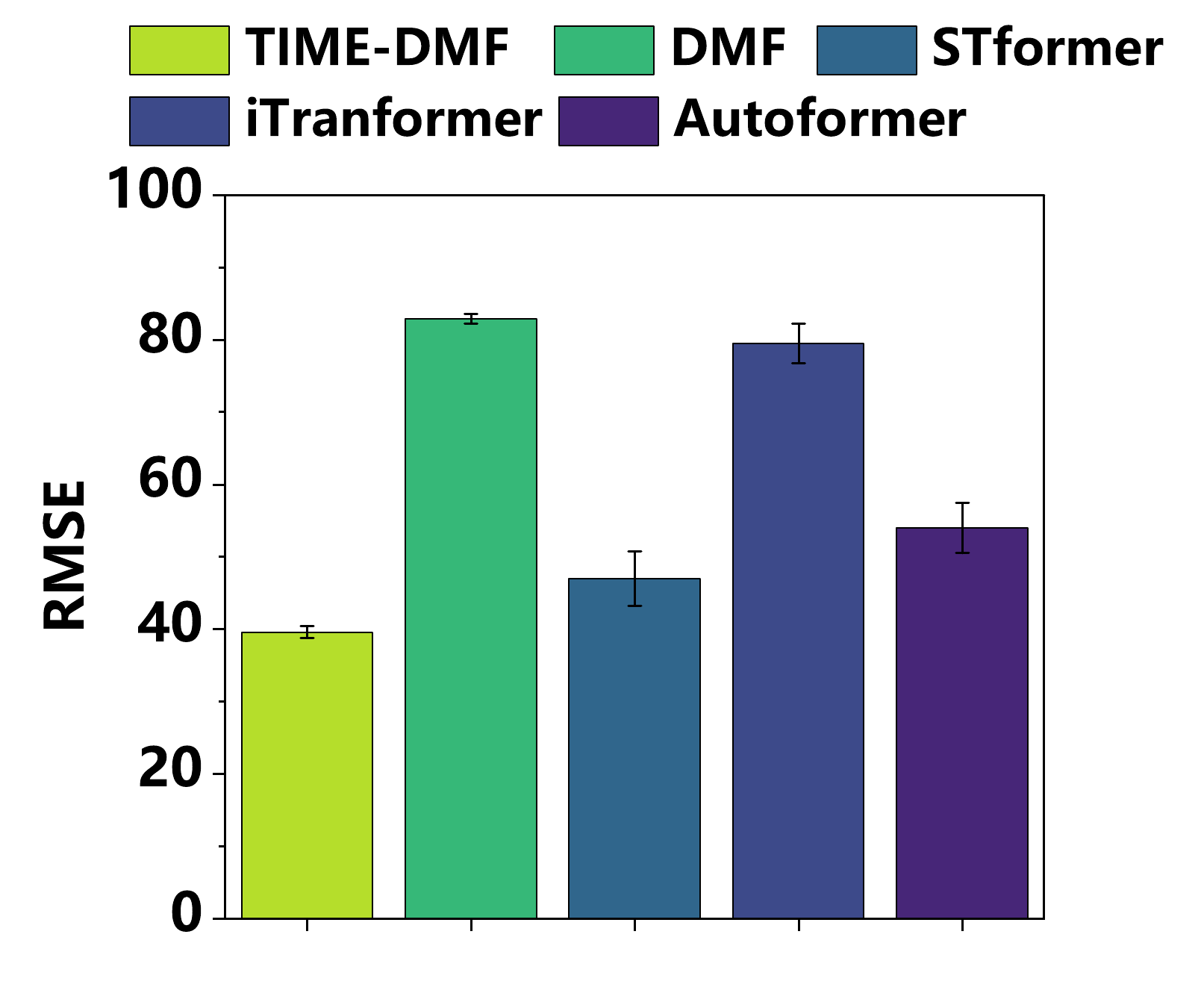}
\caption*{(b) HE-2months}
\end{minipage}  
\begin{minipage}[t]{.32\columnwidth}
\centering
\includegraphics[width=3.3cm]{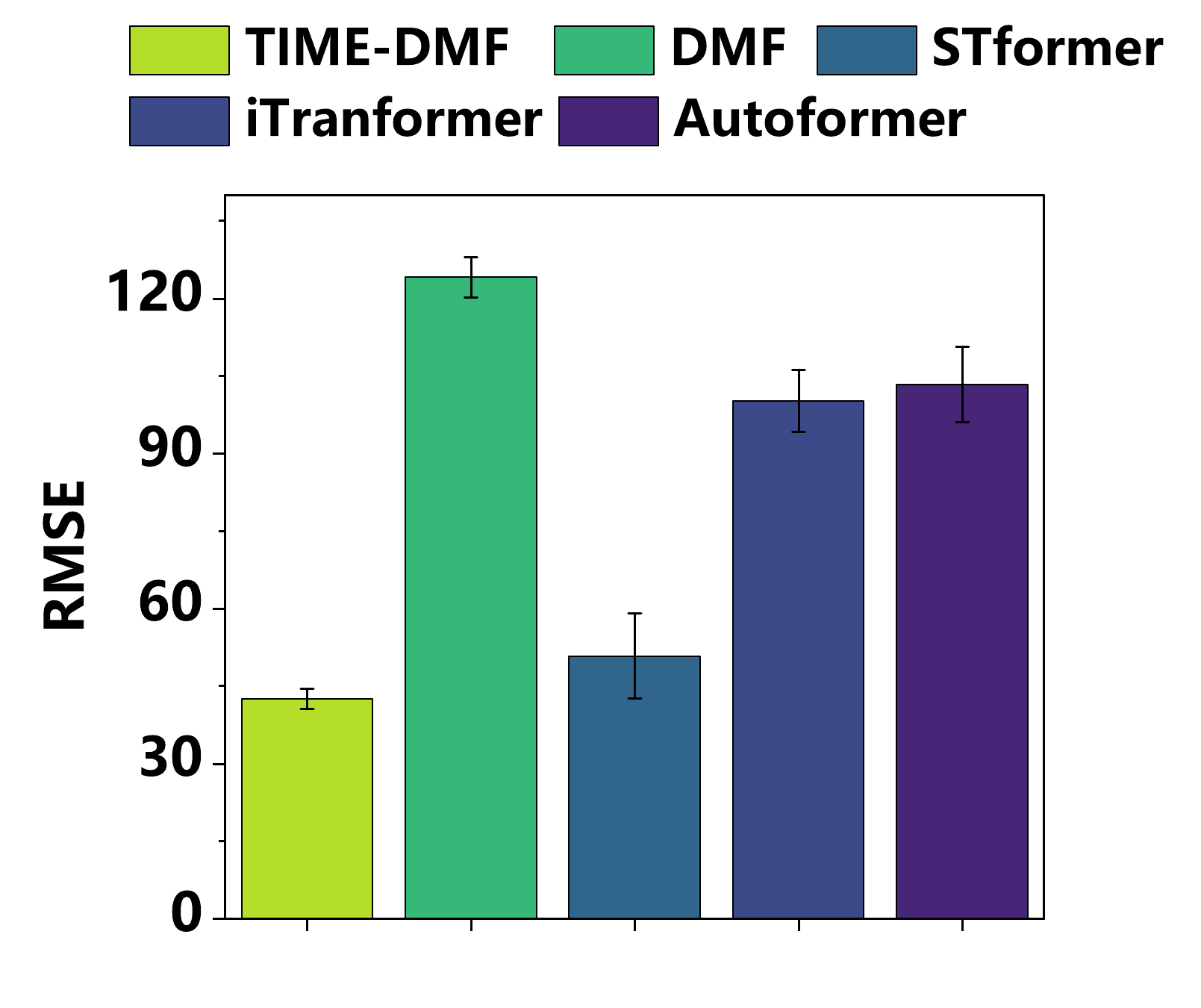}
\caption*{(c) HE-3months}
\end{minipage}
\caption{RMSE for large datasets.}
\label{fig_exp3.2}
\end{minipage}

\end{figure*}
 Since we account for data continuity, we cannot provide states for all moments. Instead, we enable our model to dynamically generate data states based on user requests. In this experiment, we test the generative capability of TIME-DMF. Note that existing time-discrete models don't have generative needs, so we compare TIME-DMF with series of predictive models. In order to make the comparison fairer, we just predict one step ahead in predictive models.

As we construct datasets via random masking and deletion, and each time we only generate data for a moment (a column in the matrix), the generation accuracy is highly unstable. For more persuasive results, we conduct extensive experiments on different randomly built datasets and depict the findings in a box plot. From the average height of the boxes in Fig. \ref{fig_exp4} we can tell that TIME-DMF provides superior overall generation accuracy. The comparison between the lengths of light green boxes and dark green boxes indicates that when the unevenness of data distribution rises, the overall performance becomes unstable. This corresponds to the conclusion that we draw in the second experiment. It should be noted that generation effect on U-AIR dataset fluctuates greatly under different deletion rates. This is because U-AIR is a rather tiny dataset, which becomes too small to contain ample temporal information as the deletion ratio increases. We do not use U-AIR in our second experiment for the same reason.

\subsection{Comparison between Time-discrete and Time-continuous completion (RQ4)}

\begin{table*}[htbp]
  \centering
  \caption{Direct comparison of time-continuous and time-discrete methods.}
  \renewcommand{\arraystretch}{1.2} 
  \setlength{\tabcolsep}{10pt}
    \begin{tabular}{c|cccccc|cc}
    \hline
    \multirow{2}{*}{Dataset} & \multicolumn{6}{c|}{Sparse-supervised Methods} & \multicolumn{2}{c}{Dense-supervised Methods} \\
    \cline{2-9}
          & GP    & KNN-S & MC    & DMF   & TIME-DMF & STformer & iTransformer & Autoformer \\
    \hline
    U-AIR & 14.0    & 40.4  & 49.7  & 50.6  & 12.6  & 13.7  & \textbf{8.8} & 10.2 \\
    Sensor-Scope & 105.7 & 93.0    & 96.9  & 103.3 & 84.8  & \textbf{80.2} & 84.4  & 90.2 \\
    TaxiSpeed & 33213.7 & 26283.4 & 23232.2 & 13129.9 & \textbf{8822.4} & 10068.7 & 10337.1 & 10748.0 \\
    Hishways England A & 66.5  & 70.5  & 66.1  & 71.5  & \textbf{22.9} & 27.3  & 47.6  & 42.6 \\
    Hishways England B & 146.7 & 129.6 & 89.1  & 82.9  & \textbf{39.6} & 47.0    & 79.5  & 54.0 \\
    Hishways England C & 207.7 & 182.0   & 109.6 & 124.1 & \textbf{42.5} & 50.8  & 100.2 & 103.4 \\
    \hline
    \end{tabular}%
  \label{table_exp3}%
\end{table*}

In time-discrete scenarios, we reduce matrix sparsity by sacrificing accurate submission times, while in time-continuous scenarios, we use time intervals but face increased sparsity. After showing that our model can handle high sparsity and effectively use intervals, we need to prove that time-continuous completion truly outperforms time-discrete solutions. To test this, we compare TIME-DMF with time-discrete methods in their original problem setting. We simulate real MCS scenarios by constructing submissions with random masking and deletion from the original datasets. For time-discrete methods, we divide the timeline into equal slices and merge data within each slice, while TIME-DMF keeps submissions in their original state with recorded arrival times. We then apply various completion methods and compare their accuracy.

After random deletion, the size of datasets will reduce a lot and completion methods may yield different results on datasets of varying sizes. To show the universality, we conduct the experiments on both smaller datasets and larger datasets. In Fig. \ref{fig_exp3.1} and Fig. \ref{fig_exp3.2}, moving from left to right, the number of columns in the datasets increases.

As shown in Table \ref{table_exp3}, Fig. \ref{fig_exp3.1} and Fig. \ref{fig_exp3.2}, TIME-DMF (which is a time-continuous completion method) significantly outperforms the traditional methods (which are all time-discrete methods).  While incorporating the continuous change of data does present new challenges such as extreme sparsity and interval handling, it is still a better model of the real scenarios if we could utilize temporal information to the maximum. When comparing the completion effects on smaller datasets and bigger ones, the performance on bigger datasets is better. This can be explained as the combination of global memories and local memories can be fully useful in the long term data distribution. While in smaller datasets, the design of global memory and local memory may seem to be unnecessary. 

Previous studies used merged data as the ground truth for model training, leading to significant deviations between experimental and real results, which caused their models to be misestimated. By evaluating the model in a way that is closer to real-world problems, we demonstrate that TIME-DMF's advanced time-continuous completion is a superior solution for real-world MCS applications.

\subsection{Model Parameter Comparison  (RQ5)}

\begin{table}[tbp]
  \centering
  \caption{Training condition, speed and parameter count.}
  \renewcommand{\arraystretch}{1.2} 
  \setlength{\tabcolsep}{2pt}
    \begin{tabular}{c|cccc}
    \hline
          & Time-DMF & iTransformer & Autoformer & STformer \\
    \hline
    Only Sparse-supervised & \checkmark   & \text{\sffamily \texttimes}   & \text{\sffamily \texttimes}   & \checkmark \\
    Parameter Count & \textbf{9k}   & 51k   & 56k   & 11M \\
    Training Time / Epoch& \textbf{18ms}  & 58ms & 66ms  & 400ms \\
    \hline
    \end{tabular}%
  \label{table:parameter}%
\end{table}

Transformer architecture has been a popular solution in time series modeling. We chose LSTM over the Transformer architecture by considering training conditions, parameter count, and speed. Fine-tuned Transformer models for completion tasks often require complete historical data for training, which is unavailable in typical MCS scenarios, making dense-supervised methods difficult to train effectively. Additionally, edge computing devices are sensitive to computational resources, so model size and training speed are critical. As shown in Table \ref{table:parameter}, TIME-DMF does not need extensive historical data and has a more compact parameter setting. Given that MCS tasks focus on specific data resources and don’t require broad generalization, we find LSTM to be a more efficient and suitable choice compared to Transformers.

\section{Conclusion}

In this paper, we propose a time-continuous completion method called TIME-DMF to challenge a widely existing assumption in Sparse MCS that data stays constant within periods and finally increase data completion accuracy to a great extent. TIME-DMF is based on  DMF which is a neural network-enabled framework for traditional data completion. Based on that, TIME-DMF is further inserted with a temporal encoder that has the function of passing temporal information between time steps and utilizing the intervals of different lengths. To pass temporal information, we imitate the structure of RNN to generate appropriate embedding vectors. For utilizing the temporal information within intervals, we design the global memory to control the overall property of data distribution and the global memory to control local trend. The length of interval serves as a parameter to monitor the memory updating process. TIME-DMF is used in cooperation with the Q-G strategy which allows users to query and then dynamically generate responses. Finally, we do extensive experiments on real-world datasets to prove the effectiveness of our models.

\newpage

\bibliographystyle{IEEEtran}
\bibliography{reference} 


 





\end{document}